\documentclass[10pt,twocolumn,letterpaper]{article}

\usepackage{iccv}
\usepackage{times}
\usepackage{epsfig}
\usepackage{graphicx}
\usepackage{amsmath}
\usepackage{amssymb}
\usepackage{wrapfig}
\usepackage{svg}
\usepackage{multirow}
\usepackage{comment}
\usepackage{import}
\usepackage{adjustbox}
\usepackage{pgfplots}
\usepackage{tikz}
\usepackage{caption}
\usepackage{subcaption}

\usepackage[ruled,vlined]{algorithm2e}
\usepackage{algorithmic}
\definecolor{commentcolor}{RGB}{110,154,155}   
\newcommand{\PyComment}[1]{\textcolor{commentcolor}{\# #1}}  
\newcommand{\PyCode}[1]{\textcolor{black}{#1}} 

\usepackage[pagebackref=true,breaklinks=true,letterpaper=true,colorlinks,bookmarks=false]{hyperref}

\iccvfinalcopy 

\def\iccvPaperID{7213} 

\begin{document}

\title{SemiISP/SemiIE: Semi-Supervised Image Signal Processor and Image Enhancement Leveraging One-to-Many Mapping sRGB-to-RAW}


\author{Masakazu Yoshimura$^1$ \quad Junji Otsuka$^1$ \quad Radu Berdan$^2$ \quad Takeshi Ohashi$^1$\\
$^1$Sony Group Corporation \quad $^2$SonyAI\\
{\tt\small \{Masakazu.Yoshimura, Junji.Otsuka, Radu.Berdan, Takeshi.a.Ohashi\}@sony.com}
}

\maketitle
\ificcvfinal\thispagestyle{empty}\fi

\begin{abstract}
   DNN-based methods have been successful in Image Signal Processor (ISP) and image enhancement (IE) tasks. However, the cost of creating training data for these tasks is considerably higher than for other tasks, making it difficult to prepare large-scale datasets. Also, creating personalized ISP and IE with minimal training data can lead to new value streams since preferred image quality varies depending on the person and use case. While semi-supervised learning could be a potential solution in such cases, it has rarely been utilized for these tasks. In this paper, we realize semi-supervised learning for ISP and IE leveraging a RAW image reconstruction (sRGB-to-RAW) method. Although existing sRGB-to-RAW methods can generate pseudo-RAW image datasets that improve the accuracy of RAW-based high-level computer vision tasks such as object detection, their quality is not sufficient for ISP and IE tasks that require precise image quality definition. Therefore, we also propose a sRGB-to-RAW method that can improve the image quality of these tasks. The proposed semi-supervised learning with the proposed sRGB-to-RAW method successfully improves the image quality of various models on various datasets.
\end{abstract}

\section{Introduction}

\begin{figure}
    \centering
    \includegraphics[width=0.475\textwidth]{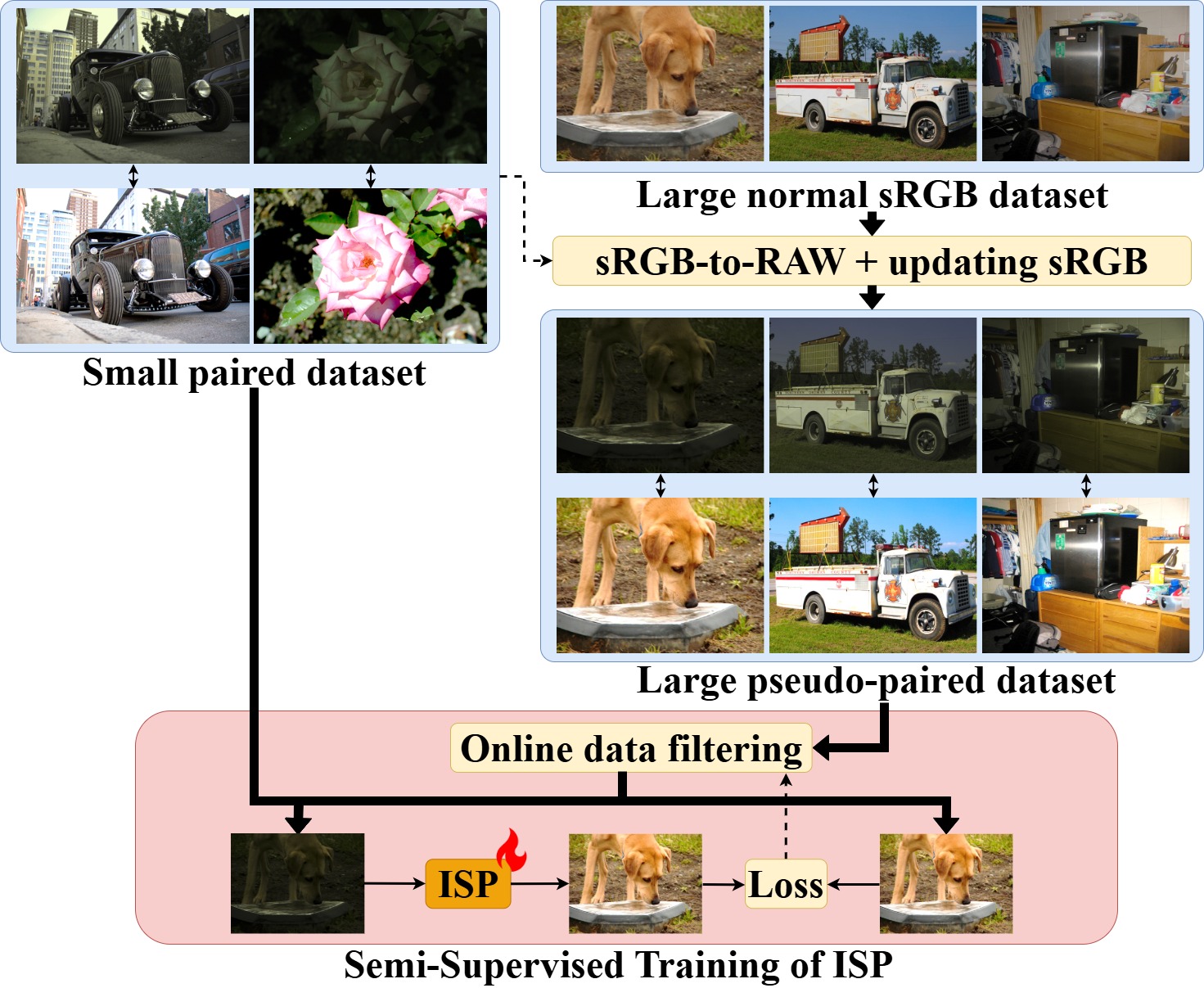}
    \vspace{-6mm}
    \caption{Overview of our proposed semi-supervised learning for ISP. We need a small number of RAW and retouched sRGB image pairs with the desired image quality. Our method generates pseudo-data using a proposed sRGB-to-RAW and a proposed method updating sRGB quality from normal quality sRGB images and improves the ISP quality with the proposed semi-supervised learning.
    }
    \label{fig:overview}
\vspace{-1mm}
\end{figure}

The Image Signal Processor (ISP) converts RAW images from image sensors into sRGB images that appear natural and pleasing to the human eye. This conversion involves various sub-tasks, including tone mapping, which compresses data distribution through non-linear transformation, and denoising, which reduces noise levels. On the other hand, image enhancement (IE) is a task that converts sRGB images captured in normal or low light conditions into higher quality sRGB images. This conversion also requires tone mapping to achieve a better luminance distribution.

In recent years, the image quality of ISP and IE methods has improved through DNN-based data-driven approaches. Many of them are trained using pairs of input and ground truth images \cite{yoshimura2024pqdynamicisp,Cai_2023_ICCV, yang2022seplut, gou2023syenet}. However, the cost of creating these paired datasets is significantly higher compared to other tasks, making it difficult to prepare large-scale datasets. For example, the LoLv1 \cite{wei2018deep} and LoLv2 \cite{yang2021sparse} datasets, which are commonly used for evaluating IE, contain only 500 and 789 images respectively. They are constructed through elaborate hardware setups. 
While these methods can be scaled once the hardware setup is established, it is challenging to capture moving subjects, such as humans. Moreover, since they use existing ISP output as ground truth, it is difficult to achieve image quality that exceeds current ISP capabilities. To achieve better image quality, it is necessary to collect expert retouched images as ground truth, such as in the FiveK dataset \cite{bychkovsky2011learning}. However, retouching requires adjusting numerous parameters with multiple trial-and-error for each image, making it extremely time-consuming. 

It is known that the definition of good image quality varies even among experts \cite{bychkovsky2011learning}. Whether natural, vivid, or moody images are suitable also varies depending on the use case. Personalized ISP and IE systems adapted to individual use cases and preferences would be valuable. Given these contexts, achieving high image quality with minimal training data is desirable.

Semi-supervised learning, which uses a combination of a large set of unlabeled data complementing a small set of labeled data, presents a potential solution to this challenge. Semi-supervised learning is successful in tasks such as classification, segmentation, and detection. Many recent methods have leveraged pseudo-labels for unlabeled data and improved the performance by refining them. For instance, MeanTeacher \cite{tarvainen2017mean}, which uses predictions from a moving average model on weakly augmented images as pseudo-labels for strongly augmented images, has gained widespread adoption. MixMatch \cite{berthelot2019mixmatch}, which applies multiple augmentations to create better pseudo-labels through consistency regularization, is also widely used. Other methods filter unlabeled data based on pseudo-label confidence \cite{feng2021ssr, xiao2023promix, rizvedefense2021}.

Applying semi-supervised learning to ISP and IE tasks presents several challenges. First, methods based on MeanTeacher or MixMatch rely on the premise that augmentations do not alter the ground truth, which does not hold for ISP and IE tasks. In addition, they are regression tasks. They lack the concept of confidence, making it impossible to employ confidence-based accuracy improvement techniques. Therefore, we propose a semi-supervised learning method designed for ISP and IE tasks, that overcomes these challenges. 

Specifically, we utilize a combination of a small set of supervised pairs consisting of input images and their desired quality target images, along with pseudo input-target pairs generated from unlabeled general quality sRGB images. We propose improving performance by estimating the quality of pseudo-data based on loss values, adapted for regression tasks.

For ISP tasks, pseudo-input images can be created using RAW image reconstruction (sRGB-to-RAW) methods, which restore RAW images from sRGB images. sRGB-to-RAW conversion has been successfully used to improve accuracy in RAW image recognition tasks, such as object detection in RAW images, by converting sRGB image object detection datasets to RAW image datasets. However, the quality of generated pseudo-RAW images has not been sufficient for ISP tasks that require precise quality definitions. We argue that this limitation stems from many existing methods using DNNs to deterministically reconstruct RAW images. As shown in Fig. \ref{fig:srgb2raw}(a), multiple RAW images are mapped to a single sRGB image via ISP. Thus, sRGB-to-RAW, the inverse problem of ISP, is a one-to-many mapping problem. Deterministic methods tend to generate biased data as shown in Fig. \ref{fig:srgb2raw}(b). Therefore, we propose a sRGB-to-RAW method that achieves high-quality one-to-many generation. Furthermore, we propose updating the sRGB image quality to match the quality defined by the small supervised dataset, thereby improving pseudo-ground-truth quality.

While input images for IE tasks are not RAW images, we achieve semi-supervised learning by adapting our proposed sRGB-to-RAW method with a simple modification to handle the difference in input formats. This enables our approach to improve image quality across various datasets for both ISP and IE tasks. 

Our key contributions are as follows:
\begin{itemize}
\item We propose a method to improve the performance of a target model by evaluating pseudo-data quality based on loss values, applicable to regression tasks.
\item We propose a sRGB-to-RAW method that aims to achieve high-quality one-to-many generation.
\item We propose a method that updates sRGB image quality to match that defined by a small supervised dataset.
\item To our knowledge, it is the first semi-supervised learning method for ISP and IE tasks that is effective across various datasets and target models.
\end{itemize}

\section{Related Works}

\subsection{ISP and Image Enhancement}

Data-driven ISP and IE techniques have achieved significant success. For instance, encoder-decoder structured DNN ISP \cite{jin2023dnf, guo2024learning, liang2021cameranet, gou2023syenet} and IE \cite{wang2023ultra, Cai_2023_ICCV, xu2022snr, bai2024retinexmamba} methods have been actively proposed. 

If filtering such as denoising is not needed, it has been found that dynamically controlling 3D lookup tables \cite{yang2022seplut, zeng2020learning, yang2022adaint, kim2024image, serrano2024namedcurves}, or classical function parameters \cite{yoshimura2024pqdynamicisp} with DNNs yields better performance than directly restoring images using DNNs. These methods have been successful in tasks where filtering is not required. On the other hand, for tasks that require denoising, methods that directly restore images with DNNs have been successful. Additionally, hybrid methods combining classical functions and DNNs have also proven effective \cite{yoshimura2024pqdynamicisp, cui2022you}. In IE, datasets like LoLv1 \cite{wei2018deep} require restoring lost information caused by quantization, and diffusion model-based methods have also been successful \cite{jiang2024lightendiffusion, hou2024global, yi2023diff}. 

In our experiments, we verify that the image quality of various ISP and IE methods can be boosted by our semi-supervised learning.

\subsection{sRGB-to-RAW}

RAW images contain more information than sRGB images and can improve image recognition performance in challenging environments \cite{yoshimura2023dynamicisp, yoshimura2023rawgment, cui2024raw, wang2025adaptiveisp}. However, the scarcity of annotated RAW images remains a problem. Therefore, research has been conducted on restoring RAW images from the abundant sRGB images. DNN-based methods have been successful, and many encoder-decoder-based methods have been proposed \cite{afifi2021cie, liu2024learnable}. CycleISP \cite{zamir2020cycleisp}, InvertibleISP \cite{xing2021invertible}, and MBISPLD \cite{conde2022model} enhance performance by implementing cycle consistency. All these methods deterministically estimate RAW images. However, since ISPs cancel out ambient light, the reverse process of sRGB-to-RAW is a one-to-many problem, as shown in Fig. \ref{fig:srgb2raw}. Recently, this ill-posed inverse problem has been mitigated by restoring images along with metadata about the shooting environment \cite{punnappurath2021spatially, li2023metadata, kim2023paramisp, zhang2024leveraging, wang2023raw}. However, these methods require metadata to be pre-stored along with the sRGB images, making them unusable for existing sRGB images. UPI \cite{brooks2019unprocessing} intentionally achieves the one-to-many relationship by randomly varying the parameters of simple inverse ISP functions. SRISP \cite{otsuka2023self} achieves a one-to-many mapping by introducing a reference RAW image input instead of metadata input. Although not solving the one-to-many relationship, RAWDiffusion \cite{reinders2024raw} has a potential to generate diverse RAW images thanks to the nature of diffusion models.

\subsection{Semi-Supervised Learning}

Semi-supervised learning has been widely studied for tasks such as image classification \cite{berthelot2019mixmatch, berthelotremixmatch, zhang2021flexmatch, chensoftmatch, nguyen2024sequencematch}, object detection \cite{xu2021end, shehzadi2024sparse, zhang2023semi}, and segmentation \cite{wang2022semi, chen2021semi, na2024switching}. One approach involves using generative models but is specialized for classification tasks. For example, some methods use the discriminator of GAN \cite{goodfellow2014generative} as an image classifier \cite{wu2019enhancing, dong2019margingan, kumar2017semi}. 
Recently, the performance of generative models has improved, leading to methods that generate training datasets such as pairs of images and segmentation labels \cite{baranchuk2021label, wu2023datasetdm, nguyen2024dataset, kupyn2024dataset, zhang2021datasetgan}. While these dataset generation methods can ideally produce paired data for ISP and IE training, generating ground truth images with fine quality definitions is challenging.

On the other hand, methods using pseudo-labels have recently been successful. For example, MeanTeacher, \cite{tarvainen2017mean} which uses the inference results of a weakly augmented image from a moving average model as pseudo-labels for a strongly augmented image, is widely used \cite{wang2022semi, chen2021semi, na2024switching}. Methods applying multiple augmentations and refining pseudo-labels through consistency regularization have also been successful \cite{berthelot2019mixmatch, berthelotremixmatch, zhang2021flexmatch, chensoftmatch, nguyen2024sequencematch}. Additionally, filtering pseudo-data based on the confidence of pseudo-labels contributes to higher accuracy \cite{feng2021ssr, xiao2023promix, rizvedefense2021}. Simple pseudo-labels can be used for ISP and IE, but as mentioned earlier, these are regression tasks, and augmentations change the ground truth, making recent pseudo-label-based methods unsuitable. 

There are only a few semi-supervised learning methods for ISP and IE. Some methods train the prediction for unsupervised data to be indistinguishable from the ground truth images by a discriminator network \cite{zhang2021real, yang2020fidelity}. 
CRNet  \cite{lee2023temporally} leverages the fact that adding noise to the input image does not change the ground truth and enhances the performance of only the denoising operations in the IE task through consistency regularization.
SMSNet \cite{malik2023semi} and LMT-GP \cite{yu2024lmt} use image quality assessment (IQA) to select good predictions or latent features from the multiple predictions and use them as pseudo ground truth data. 

\section{Method}

\begin{figure*}
    \centering
    \includegraphics[width=\textwidth]{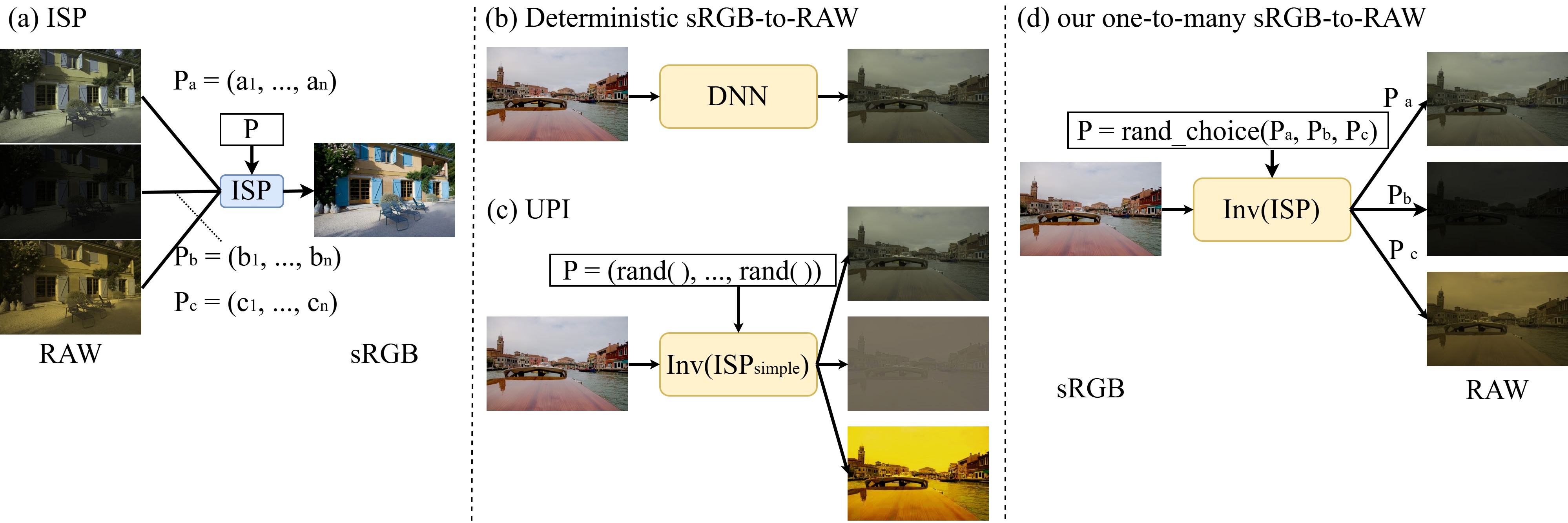}
    \vspace{-7mm}
    \caption{(a) The ISP cancels out ambient light, resulting in a many-to-one mapping. (b) Many DNN-based sRGB-to-RAW methods deterministically restore RAW images, generating realistic but biased data towards for example normal environments. (c) UPI \cite{brooks2019unprocessing} achieves one-to-many mapping by randomly assigning hyperparameters of a simple inverse ISP, but some of the outputs are unrealistic. (d) We achieve realistic quality and distribution one-to-many mapping by using the parameter set $P$ used in a high-performance ISP in its inverse function.
    }
    \label{fig:srgb2raw}
    \vspace{-1mm}
\end{figure*}

As shown in Fig. \ref{fig:overview}, we propose semi-supervised learning for ISP and IE. Below, we first describe semi-supervised learning for ISP. Then, we describe how to apply it to IE. Our method requires a small number of paired RAW images $I_R$ and target quality sRGB images $I_S$. By converting easily available normal quality sRGB images $\bar{I_S}$, we create pseudo-RAW $\tilde{I_R}$ and pseudo-sRGB $\tilde{I_S}$ paired datasets to perform semi-supervised learning. There are three main proposals for our semi-supervised learning for ISP. The first is a one-to-many mapping sRGB-to-RAW method, which generates high-quality pseudo-RAW images with realistic distributions from sRGB images. The second is a method that updates the quality of normal sRGB images to match the quality of a small number of ground-truth sRGB images, creating pseudo-ground-truth images. The third is a semi-supervised learning method that utilizes the pseudo-RAW images and pseudo-ground-truth images through loss-based online data filtering.

\subsection{One-to-Many Mapping sRGB-to-RAW}

The human eye tends to cancel out the color and brightness of ambient light, and the ISP needs to similarly cancel out ambient light to preserve the same appearance recognized by the human eye. Therefore, the ISP is a many-to-one mapping as shown in Fig. \ref{fig:srgb2raw}(a), and its inverse, sRGB-to-RAW, is an ill-posed one-to-many mapping. While many DNN-based sRGB-to-RAW methods train to restore RAW images from sRGB images, we train the opposite direction with the same dataset to avoid the ill-posed problem. In other words, we train an ISP generating sRGB images from RAW images and use its inverse function to achieve high-quality sRGB-to-RAW with a realistic data distribution.
Since it is necessary to obtain the inverse function of the ISP, we use PQDynamicISP \cite{yoshimura2024pqdynamicisp}, which utilizes a simple function-based ISP while achieves state-of-the-art accuracy by dynamically controlling the parameters of the ISP. This method determines the optimal ISP parameters $P$ for each RAW image $I_R$ using a DNN encoder as follows: 
\begin{equation}
P = DNN(Resize(I_R)) ,
\label{eq:dnn}
\end{equation}
\begin{equation}
I_S = f_{ISP}(I_R; P).
\label{eq:isp}
\end{equation}

The $I_S$ is the obtained sRGB image. Here, since the noise level in general sRGB images is low and negligible for our pseudo-data generation, we do not include filtering processes such as denoisers in the ISP. Following PQDynamicISP, we sequentially apply color correction $f_{CC}$, gain $f_{GA}$, gamma tone mapping $f_{GM}$, and contrast stretcher functions $f_{CS}$ as follows: 
\begin{equation}
f_{ISP}(I_R; P) = f_{CS}\circ f_{GM}\circ f_{GA}\circ f_{CC}(I_R; P).
\label{eq:isp2}
\end{equation}
For example, the color correction function $f_{CC}$ is formulated with nine parameters $P_{CC}$ as follows:
\begin{equation}
f_{CC}(I; P_{CC}) = \begin{bmatrix}
   p_{rr} & p_{rg} & p_{rb} \\
   p_{gr} & p_{gg} & p_{gb} \\
   p_{br} & p_{bg} & p_{bb}
\end{bmatrix}I.
\label{eq:ccm}
\end{equation}
Please refer to the Appendix for other functions.
We train the DNN encoder to control the parameters $P$ of this ISP functions with a small paired dataset using mean square loss and perceptual loss  \cite{johnson2016perceptual} with VGG16 \cite{simonyan2014very} as follows: 
\begin{equation}
L_{ISP} = L_{mse}+\alpha L_{vgg}.
\label{eq:loss_isp}
\end{equation}
We then save the ISP parameter sets $P_1$ to $P_N$ used in the last $N$ data samples during training.

The inverse function of this ISP is as follows: 
\begin{equation}
f_{ISP}^{-1}(I_S; P) = f^{-1}_{CC}\circ f^{-1}_{GA}\circ f^{-1}_{GM}\circ f^{-1}_{CS}(I_S; P).
\label{eq:inv_isp}
\end{equation}
Since the ISP is function-based, all inverse functions, except for $f_{GM}$, in the ISP are explicitly determined. 
For example the inverse function of $f_{CC}$ is as follows:
\begin{equation}
f^{-1}_{CC}(I; P_{CC}) = \begin{bmatrix}
   p_{rr} & p_{rg} & p_{rb} \\
   p_{gr} & p_{gg} & p_{gb} \\
   p_{br} & p_{bg} & p_{bb}
\end{bmatrix}^{-1}I.
\label{eq:inv_ccm}
\end{equation}

The $f_{GM}$ is parameterized by three parameters $P_{GM}^{(c)}=\{{p_{g1}^{(c)}, p_{g2}^{(c)}, p_{k}^{(c)}}\}$ for each color channel $c$ and is not invertible. Therefore, we approximate its inverse function using a 4-D lookup table $T$ that adds one dimension for the input pixel to these three parameters. Verification has shown that linear interpolation of the 4-D lookup table is sufficient to achieve an accurate inverse function without a speed bottleneck. Therefore, the inverse function of $f_{GM}$ is expressed as
\begin{equation}
\resizebox{0.908\hsize}{!}{$
f_{GM}^{(c)-1}(i^{(c)}; P_{GM}^{(c)})=interpolate(T,(i^{(c)},p_{g1}^{(c)},p_{g2}^{(c)},p_{k}^{(c)})),
\label{eq:lut}
$}
\end{equation}
for each pixel $i^{(c)} \in I$.
More details on how to obtain the lookup table $T$, the error of the inverse function, and the speed, please refer to the Appendix.

To achieve one-to-many sRGB-to-RAW, we randomly select from the saved parameter sets $P_1$ to $P_N$ and input them into $f_{ISP}^{-1}$. In other words, our sRGB-to-RAW inference is as follows:
\begin{equation}
f_{ISP}^{-1}(\bar{I_S}; P_i)\; (i=Uniform(\{1, 2, ..., N\})).
\label{eq:srgb2raw}
\end{equation}

The major differences in ISP parameters are due to ambient light, so if $P_1$ is a parameter set used for a dark environment, $f_{ISP}^{-1}(I_S; P_1)$ will convert $I_S$ to a RAW image as if it were taken in a dark environment.

\begin{figure*}
    \centering
    \includegraphics[width=\textwidth]{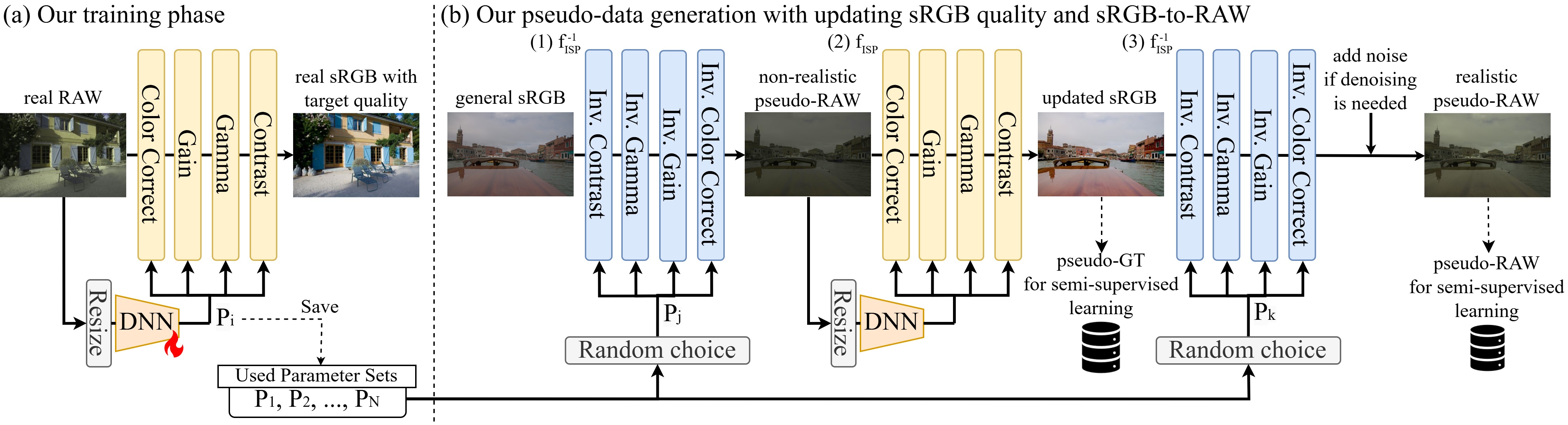}
    \vspace{-6mm}
    \caption{(a) In the training phase of our sRGB-to-RAW, we train the ISP direction, especially the DNN part to generate ISP parameters $P$. (b) In the pseudo-data generation phase, we apply our sRGB-to-RAW and ISP to update the quality of the general sRGB images similar to that of the real sRGB images. Then, we further apply our sRGB-to-RAW to generate realistic pseudo-RAW images.
    }
    \label{fig:update}
\end{figure*}

\subsection{Updating sRGB quality}

Existing sRGB-to-RAW methods are trained on target paired datasets from a specific sensor and then used to generate pseudo-RAW images from general unpaired sRGB images. However, if the image quality of the sRGB images in the paired and unpaired datasets differs, the trained method is not optimal for the unpaired sRGB images, resulting in unrealistic pseudo-RAW images. Moreover, sRGB images with different quality cannot serve adequately as a ground truth for ISP. Therefore, we propose a method to update the image quality of unpaired sRGB images to match that of the paired dataset. This enables the generation of more realistic pseudo-RAW and pseudo-ground truth pairs. The challenge is that the sRGB images in the paired and unpaired datasets are unpaired, which is why existing methods use the unpaired dataset's sRGB images as they are.

In our sRGB-to-RAW training, we train the ISP, not sRGB-to-RAW. We use the learned ISP to address this issue. First, we perform pseudo-RAW conversion similar to conventional methods, as shown in Fig. \ref{fig:update}(b)(1). Due to differing input sRGB quality, the generated pseudo-RAW is not realistic. Next, we apply the learned ISP, deterministically deciding the ISP parameter set using the DNN encoder, as shown in Fig.\ref{fig:update}(b)(2).
This DNN is trained to convert the image to the quality defined by the paired dataset, making the generated sRGB image closer to the paired dataset's quality. Then, by applying sRGB-to-RAW again, we generate more realistic pseudo-RAW images, as shown in Fig. \ref{fig:update}(b)(3). Additionally, the updated sRGB images are used as pseudo-ground truth images for semi-supervised learning.

During experiments, we found that the quality of pseudo-RAW data should be adjusted on the basis of the amount of training data. With limited training data, we should generate realistic pseudo-RAW data matching real data distribution. With sufficient training data, it is better to pre-train with diverse pseudo-data, even if somewhat unrealistic.
Therefore, when there are sufficient training data, we randomly select the ISP parameter set not from ISP unit, i.e.  \{$P_1$, $P_2$ , ... $P_{N}$\}, but from ISP function units, such as  \{$P_{CC,1}$, $P_{CC,2}$ , ... $P_{CC,N}$\} and  \{$P_{GM,1}$, $P_{GM,2}$ , ... $P_{GM,N}$\}, to cover a broader domain than real RAW images. 

\subsection{Semi-Supervised Learning}

As shown in Fig. \ref{fig:overview}, we try to achieve a higher quality ISP $F_{ISP}$ by adding the pseudo-ground truth images and pseudo-RAW images generated by the above methods to the real paired dataset. Note that $F_{ISP}$ can be any learnable ISP other than $f_{ISP}$. A mini-batch is constructed with $N_r$ real samples from the paired dataset and $N_p$ pseudo samples from the pseudo-dataset. In tasks such as classification, segmentation, and detection, where confidence predictions exist, the quality of pseudo-data has been determined based on predicted confidence values \cite{feng2021ssr, xiao2023promix, rizvedefense2021}. However, this is not possible for the ISP regression task. Therefore, we propose a simple online pseudo-data filtering based on loss values. Specifically, data with a loss value higher than that of real pairs are considered an outlier and excluded from the loss calculation. If the average of the loss value for real samples is
\begin{equation}
L_r = \frac{1}{N_r}\sum^{N_r}_{n=1}L(F_{ISP}(I_R^{(n)},),I_S^{(n)}),
\label{eq:loss_real}
\end{equation}
we filter the pseudo-data as follows:
\begin{equation}
L_p = \frac{1}{\sum^{N_p}_{n=1}\delta_n} \sum^{N_p}_{n=1}\delta_n L(F_{ISP}(\tilde{I_R^{(n)}}),\tilde{I_S^{(n)}}),
\label{eq:loss_pseudo}
\end{equation}
where $\delta_n$ is a binary mask based on a threshold $\beta$:
\begin{equation}
\delta_n = \begin{cases} 
0 & \text{if } L(F_{ISP}(\tilde{I_R^{(n)}}),\tilde{I_S^{(n)}}) > \beta L_{r} \\
1 & \text{else }
\end{cases}.
\label{eq:loss_thresh}
\end{equation}
This data filtering is adaptively applied based on the loss values of the real data at that time, allowing the filtering of pseudo-data suitable for the model at that time. Here, the amount of data actually used in the loss calculation changes with each iteration due to this filtering. If the scale of the loss changes, it can adversely affect the momentum information of the gradients within optimizers. Therefore, the balance of the loss is scaled according to the amount of data as follows:
\begin{equation}
L_{semi} = \frac{N_r}{N_r+\sum^{N_p}_{n=1}\delta_n}L_r+\frac{\sum^{N_p}_{n=1}\delta_n}{N_r+\sum^{N_p}_{n=1}\delta_n}L_p.
\label{eq:loss_semi}
\end{equation}
After this training, we perform supervised fine-tuning using only the paired data. 

\subsection{Application to Tasks Including Denoising}

To make the ISP function invertible, a denoiser that is hard to reverse was excluded from $f_{ISP}$. However, a denoiser is needed in some cases. In such cases, the following approach is used. If the RAW images in the small paired data $I_R$ have noise, the trained tone mapping with eq. \ref{eq:loss_isp} will be incomplete due to noise outliers. Therefore, the noise in the RAW images is reduced in advance using a pre-trained denoiser. Since perfect denoising is difficult, we use L1 loss, which is less affected by outliers, along with average pooling $A_k$ with kernel size $k$ to reduce noise impact as follows: 
\begin{equation}
\resizebox{0.85\hsize}{!}{$
L_{ISP}(X,Y) = L_1(X,Y)+\sum_{k}L_{1}(A_k(X),A_k(Y)).
\label{eq:loss_for_noisy}
$}
\end{equation}
This reduces the noise influence in the RAW images of the paired data, allowing proper tone mapping to be learned.

When generating pseudo-data, we generate pseudo-RAW images from normal noise-less images, and Gaussian noise is added to the pseudo-RAW images as follows to create noisy input images, allowing $F_{ISP}$ to learn to denoise:
\begin{equation}
\tilde{I_R'}= \mathcal{N}(\tilde{I_R}, \sigma_s^2\tilde{I_R}+\sigma^2_r),
\label{eq:add_noise}
\end{equation}
where $\sigma_s^2$ and $\sigma_r^2$ are random parameters per image.

\subsection{Application to Image Enhancement}

The IE task involves converting dark or noisy sRGB images into bright and high-quality sRGB images. Although the input is not a RAW image, it is a many-to-one mapping where various levels of darkness are mapped to a single appropriate brightness level, similar to ISP. Therefore, sRGB-to-degraded-sRGB conversion for semi-supervised learning in the IE task should also be a one-to-many mapping. This can be addressed simply by adding an inverse tone mapping function at the beginning of the $f_{ISP}$ in eq. \ref{eq:isp2} to handle the difference in input formats between the RAW and sRGB images, similar to PQDynamicISP.

\section{Experiments}

\begin{table*}[t]
\centering
\caption{Evaluation on the FiveK tone mapping task and FiveK general ISP task with 100, 500, and 4500 paired training images. In SemiISP (ours), sRGB images from COCO dataset are used as normal sRGB images.}
\vspace{-2mm}
\scalebox{0.91}{

\label{tab:fivek}

\begin{tabular}{lcccccc|cccccc}
\hline 
 & \multicolumn{6}{c}{FiveK tone mapping}& \multicolumn{6}{c}{FiveK general ISP}\\
\hline 
 \# images & \multicolumn{2}{c}{100} & \multicolumn{2}{c}{500} & \multicolumn{2}{c}{4500 (full)} &  \multicolumn{2}{c}{100} & \multicolumn{2}{c}{500} & \multicolumn{2}{c}{4500 (full)}
\\
 \hline 
 & PSNR & SSIM & PSNR & SSIM & PSNR & SSIM  & PSNR & SSIM & PSNR & SSIM & PSNR &SSIM  
\\
\hline 
\hline 
MicroISP \cite{ignatov2022microisp} & - & - & - & - & 24.07 & 0.909 
 & - & - & - & - & 21.64&0.885\\
HDRNet \cite{gharbi2017deep} & - & - & - & - & 24.52 & 0.915 
 &  
- & - & - & - & -&-\\
3D LUT \cite{zeng2020learning} & - & - & - & - & 25.06 & 0.920 
 & - & - & - & - & -&-\\
SYENet \cite{gou2023syenet} & - & - & - & - & 25.19 & 0.922 
 & - & - & - & - & 22.24&0.889\\
AdaInt \cite{yang2022adaint} & - & - & - & - & 25.28 & 0.925 
 &  
- & - & - & - & - &- \\
F. Zhang \etal \cite{zhang2024lookup} & - & - & - & - & 25.53 & 0.907 
 & - & - & - & - & - &- \\
LUTwithBGrid \cite{kim2024image} & - & - & - & - & 25.59 & 0.932 
 &  
- & - & - & - & - &- \\
\hline 
NAFNet-s \cite{chen2022simple} & 22.73 & 0.889 & 23.50& 0.886& 24.73 & \textbf{0.920}& 19.39& 0.777& 20.27& 0.800& 22.53&0.876\\
 \quad+ SemiISP (ours)& \textbf{23.81}& \textbf{0.907}& \textbf{24.20}& \textbf{0.917}& \textbf{24.87}&  0.918&  
\textbf{21.48}& \textbf{0.872}& \textbf{22.01}& \textbf{0.874}& \textbf{22.55}&\textbf{0.880}\\
SepLUT \cite{yang2022seplut} & 21.54& 0.858& 22.38& 0.871& 25.43 & \textbf{0.922} & 20.11& 0.828& 20.84& 0.843& 22.82&0.891\\
 \quad+ SemiISP (ours)& \textbf{23.98}& \textbf{0.905}& \textbf{24.67}& \textbf{0.912}& \textbf{25.45}& \textbf{0.922} &  
\textbf{21.66}& \textbf{0.873}& \textbf{22.12}& \textbf{0.878}& \textbf{23.18}&\textbf{0.897}\\
PQDynamicISP \cite{yoshimura2024pqdynamicisp} & 23.07 & 0.898& 23.50& 0.900& 25.72& 0.933 & 20.62& 0.864& 20.90& 0.865& 23.59&0.911\\
 \quad+ SemiISP (ours)& \textbf{24.12}& \textbf{0.910}& \textbf{24.43}& \textbf{0.910}& \textbf{25.73}& \textbf{0.934} &  
\textbf{21.94}& \textbf{0.885}& \textbf{22.11}& \textbf{0.886}& \textbf{23.61}&\textbf{0.913}\\
\hline 
\end{tabular}
}
\vspace{-2mm}
\end{table*}

\subsection{Datasets and Tasks}
We use the FiveK dataset \cite{bychkovsky2011learning} for the ISP task, and the LoLv1 \cite{wei2018deep}, LoLv2 \cite{yang2021sparse}, and LoLv2-syn \cite{yang2021sparse} datasets for the IE task. LoLv1 and LoLv2 need denoising because the input images are noisy. In all tasks, we use sRGB images from COCO dataset \cite{lin2014microsoft} as normal sRGB images.

\textbf{FiveK tone mapping task:}
FiveK contains RAW images from 35 different image sensors paired with sRGB images from hardware ISPs and manually retouched ground-truth sRGB images. Many works \cite{chen2018deep,zeng2020learning, yoshimura2024pqdynamicisp} pre-convert the RAW images into CIE XYZ images canceling sensor characteristics using camera metadata, and then predict sRGB images retouched by expert C. This is a so-called tone mapping task, and we also follow this convention but treat it as an ISP task. It is because, although the CIE XYZ images have already undergone color correction and white balance, these are incomplete, and additional color correction and white balance are necessary. 4500 pairs are used as training data and 500 pairs as test data. Additional experiments are conducted using subsets of training data consisting of only 500 and 100 images. The 480P-sized images are used. 

\textbf{FiveK general ISP task:}
Unlike the FiveK tone mapping task, this task converts RAW images into the retouched sRGB images. This is a challenging task as it deals with various image sensors without using camera metadata. Following PQDynamicISP, only demosaicing is performed in advance, and the evaluation is conducted on 480P-sized images.


\textbf{LoLv1, LoLv2, and LoLv2-syn IE task:}
These tasks convert low light sRGB images into well-lit clean sRGB images. The training data consist of 485 and 689, and 900 pairs, respectively.

\subsection{Implementation Details}

Our method consists of three training phases. The first phase involves training $f_{ISP}$ to generate pseudo-data. The second phase is semi-supervised learning of the target model $F_{ISP}$. The final phase is the fine-tuning of $F_{ISP}$ using only supervised data. We use the same hyperparameters as possible but adjust them for the first two training phases based on the amount of training data and whether a denoising is needed. During fine-tuning, we use the hyperparameters from the original paper of the target model. Since some training has already been done, we halves the number of iterations compared to the original paper. The details are provided in the Appendix.

There are two types in PQDynamicISP \cite{yoshimura2024pqdynamicisp}, which we use as $f_{ISP}$: one that controls ISP parameters globally for each image and one that controls locally for coarse regions. Although local control offers higher quality, our $f_{ISP}^{-1}$ uses a random parameter set regardless of the input image, which undesirably enhance certain local areas. Therefore, we use the global one for $f_{ISP}^{-1}$ in Fig. \ref{fig:update} (b)(1). 
On the other hand, since $f_{ISP}$ in Fig. \ref{fig:update} (b)(2) is a deterministic process that decides the parameters based on the input image, we use high-performance local control one. As to $f_{ISP}^{-1}$ in Fig. \ref{fig:update} (b)(3), as we mentioned earlier, we generate realistic pseudo-RAW images with global one if the training data are limited (less than 1,000 images). On the other hand, we use local one in addition to random parameter choice per ISP function units as augmentation. This allows the model to learn to cancel out non-uniform ambient light.





\begin{table}[t]
\centering
\caption{Ablation studies of the proposed online pseudo-data filtering (OPF) and updating sRGB quality (USQ) on FiveK tone mapping. The target model is PQDynamicISP.}

\scalebox{0.78}{

\label{tab:ablation}

\begin{tabular}{ccccccc}
\hline 
 & & \# images & \multicolumn{2}{c}{100}  & \multicolumn{2}{c}{4500 (full)}\\
 \hline 
sRGB-to-RAW & OPF& USQ& PSNR & SSIM & PSNR & SSIM\\
\hline 
\hline 
UPI &  & & 22.79& 0.889& 25.44& 0.930\\
UPI &  \checkmark & \checkmark & \textbf{23.40}& \textbf{0.904}& \textbf{25.67}& \textbf{0.932}\\
\hline 
ReRAW&  & & 23.15& 0.900& \textbf{25.57}& \textbf{0.931}\\
ReRAW&  \checkmark & \checkmark & \textbf{23.82}& \textbf{0.901}& \textbf{25.57}& 0.930\\
\hline 
ours &  & & 23.07& 0.890& 25.57& 0.930\\
 ours & \checkmark & & 23.31& 0.891& 25.63&0.930\\
ours &  \checkmark & \checkmark & \textbf{24.12}& \textbf{0.910}& \textbf{25.73}& \textbf{0.934}\\
\hline 
\end{tabular}
}
\end{table}

\begin{figure*}
\centering
\renewcommand{\arraystretch}{0.2}
\setlength{\tabcolsep}{0.5pt}
\scalebox{0.646}{
\begin{tabular}{cccccccc|ccc}
\includegraphics[width=0.13785\textwidth]{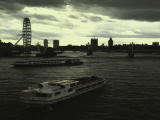} & \includegraphics[width=0.13785\textwidth]{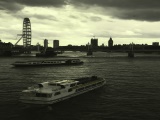}  &\includegraphics[width=0.13785\textwidth]{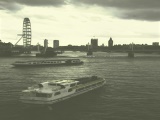}
&
\includegraphics[width=0.13785\textwidth]{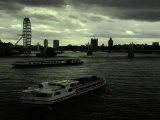}& \includegraphics[width=0.13785\textwidth]{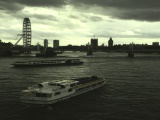}&\includegraphics[width=0.13785\textwidth]{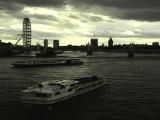} &\includegraphics[width=0.13785\textwidth]{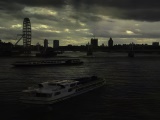}& \includegraphics[width=0.13785\textwidth]{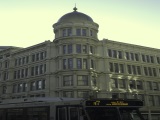}&
\includegraphics[width=0.13785\textwidth]{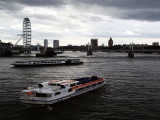}&
\includegraphics[width=0.13785\textwidth]{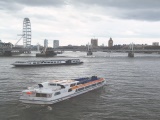}&
\includegraphics[width=0.13785\textwidth]{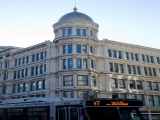}\\
 &  &\includegraphics[width=0.13785\textwidth]{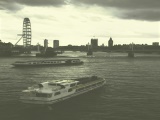}
& \includegraphics[width=0.13785\textwidth]{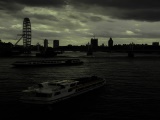}& \includegraphics[width=0.13785\textwidth]{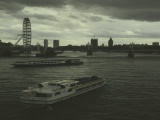}&\includegraphics[width=0.13785\textwidth]{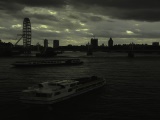} &\includegraphics[width=0.13785\textwidth]{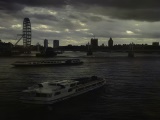}& \includegraphics[width=0.13785\textwidth]{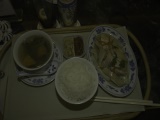}& \includegraphics[width=0.13785\textwidth]{figs/viz_srgb2raw/000000008277_ori}& \includegraphics[width=0.13785\textwidth]{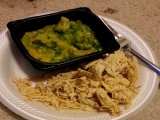}&\includegraphics[width=0.13785\textwidth]{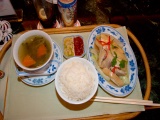}\\
 &  &\includegraphics[width=0.13785\textwidth]{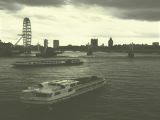}
& \includegraphics[width=0.13785\textwidth]{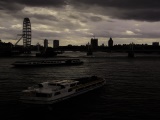}& \includegraphics[width=0.13785\textwidth]{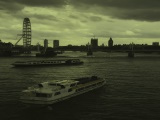}&\includegraphics[width=0.13785\textwidth]{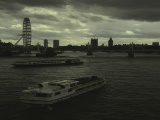} &\includegraphics[width=0.13785\textwidth]{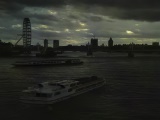}& \includegraphics[width=0.13785\textwidth]{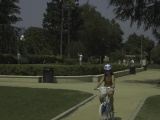}& \includegraphics[width=0.13785\textwidth]{figs/viz_srgb2raw/000000255664_ori.jpg}& \includegraphics[width=0.13785\textwidth]{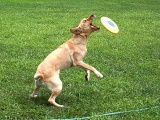}&\includegraphics[width=0.13785\textwidth]{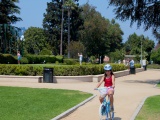}\\
\large{CycleISP} & \large{ReRAW}  &RAWDiffusion & \large{UPI} & \large{SRISP} &\large{ours} &\large{ours*} & \large{FiveK}& \large{COCO}& \large{updated}& \large{FiveK}\\
\end{tabular}
}
\vspace{-3mm}

\caption{The left is the visual comparison of the sRGB-to-RAW methods on FiveK tone mapping task with 100 training samples. Three variations are shown for non-deterministic sRGB-to-RAW methods. Ours* is used when there is sufficient training data to cover a slightly wider domain than real RAW images. On the right is an example where normal and various quality sRGB images from the COCO dataset are converted to sRGB images in the style of FiveK's ground truth images using the proposed method. }
\vspace{0mm}
\label{fig:viz}
\end{figure*}

\begin{table}[t]
\centering
\caption{Ablation studies on sRGB-to-RAW methods on FiveK tone mapping task with PQDynamicISP as a target model.  To fairly evaluate the sRGB-to-RAW methods themselves, we use the proposed updating sRGB and loss value-based data filtering for all methods. }
\vspace{-2mm}
\scalebox{0.91}{

\label{tab:srgb2raw}

\begin{tabular}{lcccc}
\hline 
\# images & \multicolumn{2}{c}{100}  & \multicolumn{2}{c}{4500 (full)}   \\
\hline 
sRGB-to-RAW & PSNR & SSIM & PSNR & SSIM \\
\hline 
\hline 
UPI \cite{brooks2019unprocessing} & 23.40& 0.904& 25.67& 0.932\\
 CycleISP \cite{zamir2020cycleisp}& 23.32& 0.900& 25.55&0.929\\
 SRISP \cite{otsuka2023self}& 23.39& 0.899& 25.62&0.932\\
 RAWDiffusion \cite{reinders2024raw}& 23.21& 0.897& 25.63&0.932\\
 ReRAW \cite{radu2025reraw}& 23.82& 0.901& 25.57&0.930\\
ours & \textbf{24.12}& \textbf{0.910}& \textbf{25.73}& \textbf{0.934}\\
\hline 
\end{tabular}
}
\vspace{-1mm}
\end{table}

\subsection{Evaluation on ISP Task}

We benchmark and perform ablation studies on FiveK tone mapping and general ISP tasks. To check the versatility of our method, we test it on three target models: NAFNet \cite{chen2022simple} (a large encoder-decoder DNN model), SepLUT \cite{yang2022seplut} (a 3D lookup table-based method), and PQDynamicISP \cite{yoshimura2024pqdynamicisp} (a simple function-based method). Table \ref{tab:fivek} shows that our method improves all models, especially with limited 100 and 500 training images. As shown in the visual comparison in the Appendix, our method realizes ISPs that are stable and high-quality with just 100 training pairs. Since preferred image quality varies by individual and use case, the proposed method, which can improve performance with limited data, can create new value, such as personalized ISPs. Additionally, even with a large amount of training data, the proposed method does not degrade the quality and further improves the quality of state-of-the-art methods.

Table \ref{tab:ablation} verifies the effects of our online pseudo-data filtering (OPF) and updating sRGB quality (USQ). Both OPF and USQ improve performance across different sRGB-to-RAW methods. OPF is particularly effective for UPI \cite{brooks2019unprocessing}, SRISP \cite{otsuka2023self}, and the proposed one-to-many sRGB-to-RAW. This is likely because these methods generate a variety of pseudo-data, some of which are not realistic. Next, Table \ref{tab:srgb2raw} compares the sRGB-to-RAW methods. For fairness, OPF and USQ are applied to all methods. 
Deterministic and high-precision sRGB-to-RAW methods, such as CycleISP \cite{zamir2020cycleisp} and ReRAW \cite{radu2025reraw}, generate realistic but less varied pseudo-data, as shown in Fig. \ref{fig:viz}. When the training data are limited, these realistic pseudo-data improve the quality of ISP, but when there is sufficient training data, the pseudo-data with biased domain degrades the quality. On the other hand, our sRGB-to-RAW, as well as UPI and SRISP, which achieve one-to-many mapping, generate pseudo-data with rich variation, and when combined with the proposed OPF, they achieve relatively high performance even with sufficient training data. Among them, our sRGB-to-RAW, which intentionally mimics real data distribution in a learning-based manner, generates more realistic data with realistic distribution, and performs well with both limited and sufficient training data. SRISP is also learning-based but it sometimes outputs unrealistic images due to the different training and inference pipeline.

\subsection{Evaluation on IE Task}

We apply our method to LLFormer \cite{wang2023ultra} and Retinexformer \cite{Cai_2023_ICCV}. For detailed experimental settings, please refer to the Appendix. Unlike the ISP task, several semi-supervised learning methods such as DRBN \cite{yang2020fidelity},  SMSNet \cite{malik2023semi}, CRNet \cite{lee2023temporally}, and LMT-GP \cite{yu2024lmt} have been proposed for the IE task, so we also compare with these methods. The results are shown in Table \ref{tab:lol}.

First, we successfully improve the performance of LLFormer and Retinexformer using the proposed semi-supervised learning, demonstrating that our one-to-many mapping sRGB-to-RAW idea can also be applied to the IE task. 
Second, our method achieves high quality compared to existing semi-supervised learning methods. Our approach is effective against adversarial loss-based DRBN and CRNet, and IQA-based SMSNet. Additionally, our method is versatile and easy to use; the target model can be any model while methods like DRBN, CRNet and LMT-GP require a specific target model structure, the additional training data are easily available normal quality sRGB images while methods like LMT-GP require pristine sRGB images.

\begin{table}[t]
\centering
\caption{Evaluation on LoLv1, LoLv2, and LoLv2-syn IE task. We report the metric after fixing global color using ground truth mean in (\;) following KinD \cite{zhang2019kindling}}
\vspace{-2mm}
\scalebox{0.79}{
\label{tab:lol}

\begin{tabular}{lccc}
 \hline 
 & LoLv1& LoLv2& LoLv2-syn\\
 & PSNR & PSNR & PSNR \\
\hline 
\hline 
 LLFlow \cite{wang2022low}& 21.15 (25.19) & 17.43 (25.42)& 24.81 (27.96)\\
 SNR-Aware \cite{xu2022snr} & 24.61 (26.72) & 21.48 (27.81)& 24.14 (27.79)\\
 CIDNet \cite{yan2025hvi}& 23.81 (27.72) & 24.11 (28.13)& 25.67 (28.99)\\
\hline 
DRBN \cite{yang2020fidelity} & 16.29 (19.55)& 20.29 \;(\;\;\;-\;\;\;)& 23.22 \;(\;\;\;-\;\;\;)\\
SMSNet \cite{malik2023semi} & 21.91 \;(\;\;\;-\;\;\;)& \;\;\;\;-\;\;\; \;(\;\;\;-\;\;\;)  
& \;\;\;\;-\;\;\; \;(\;\;\;-\;\;\;)  
\\
CRNet \cite{lee2023temporally} & \;\;\;\;-\;\;\; \;(24.01) & \;\;\;\;-\;\;\; \;(\;\;\;-\;\;\;)  
& \;\;\;\;-\;\;\; \;(\;\;\;-\;\;\;)  
\\
LMT-GP \cite{yu2024lmt} & 24.12 \;(\;\;\;-\;\;\;) & \;\;\;\;-\;\;\; \;(\;\;\;-\;\;\;)  & \;\;\;\;-\;\;\; \;(\;\;\;-\;\;\;)  \\
\hline 
LLFormer \cite{wang2023ultra} & 23.65 (25.76) & 20.06 (26.20)& 24.04 (28.01)\\
\quad+ SemiIE (ours) & \textbf{23.77 (26.12)}& \textbf{21.66 (28.08)}& \textbf{25.33 (28.79)}\\
Retinexformer \cite{Cai_2023_ICCV} & \textbf{23.89 (27.70)}& 21.81 (28.16)& 25.89 (29.37)\\
\quad+ SemiIE (ours) &  23.78 (\textbf{27.00})& \textbf{22.34 (28.39)}& \textbf{26.17 (29.40)}\\
\hline 
\end{tabular}
}
\vspace{-1mm}
\end{table}


\section{Conclusion}

We achieve semi-supervised learning for ISP and IE with three main proposals. One is the one-to-many mapping sRGB-to-RAW, which generates pseudo-RAW images with realistic quality and distribution. It is achieved by training with the ISP direction instead of the ill-posed sRGB-to-RAW direction. The second is updating sRGB quality to be similar to that of the paired dataset. We achieve it uniquely without preparing paired source sRGB and target sRGB images by applying the sRGB-to-RAW and ISP sequentially. The last is online pseudo-data filtering, which is simple but effective and usable for regression tasks. Our semi-supervised learning achieves high quality even with limited paired dataset. It can be used to easily create ISP and IE with personalized image quality for each individual and use case.

{\small
\bibliographystyle{ieee_fullname}
\bibliography{main_review}
}


\clearpage
\setcounter{page}{1}
\maketitlesupplementary

\appendix

\section{Implementation Details}

\subsection{ISP functions}

We use identical ISP functions with PQDynamicISP \cite{yoshimura2024pqdynamicisp}. In the following, we show the functions and their inverse functions for our sRGB-to-RAW. As mentioned in the main paper, we sequentially apply color correction $f_{CC}$, gain $f_{GA}$, gamma tone mapping $f_{GM}$, and contrast stretcher $f_{CS}$ functions for ISP tasks. For IE tasks, we add an inverse tone mapping function $f_{IT}$ at the beginning of the ISP.

\subsubsection*{Color Correction}

As mentioned in the main paper, the color correction function $f_{CC}$ consists of nine controllable parameters $P_{CC}$ as follows:
\begin{equation}
f_{CC}(I; P_{CC}) = \begin{bmatrix}
   p_{rr} & p_{rg} & p_{rb} \\
   p_{gr} & p_{gg} & p_{gb} \\
   p_{br} & p_{bg} & p_{bb}
\end{bmatrix}I.
\label{eq:ccm_appendix}
\end{equation}
The inverse function of $f_{CC}$ is 
\begin{equation}
f^{-1}_{CC}(I; P_{CC}) = \begin{bmatrix}
   p_{rr} & p_{rg} & p_{rb} \\
   p_{gr} & p_{gg} & p_{gb} \\
   p_{br} & p_{bg} & p_{bb}
\end{bmatrix}^{-1}I.
\label{eq:inv_ccm_appendix}
\end{equation}

\subsubsection*{Gain}

The gain function $f_{GA}$ consists of three controllable parameters $P_{GA}^{(c)}=\{{p_{x}^{(c)}, p_{w}^{(c)}, p_{h}^{(c)}}\}$ for each color channel, $c=\{r,g,b\}$, as follows:
\begin{equation}
\resizebox{1.0\hsize}{!}{$
f_{GA}(I^{(c)}; P_{GA}^{(c)})=\begin{cases}
\frac{1-p_{h}^{(c)}}{1-p_{w}^{(c)}}i^{(c)}\;\;\;\;\left(\text{if}\,i^{(c)}<p_{x}^{(c)}\left(1-p_{w}^{(c)}\right),\,i^{(c)}\in I^{(c)}\right)\\
\frac{1-p_{h}^{(c)}}{1-p_{w}^{(c)}}i^{(c)}+\frac{p_{h}^{(c)}-p_{w}^{(c)}}{1-p_{w}^{(c)}}\;\;\;\,\left(\text{if}\,p_{x}^{(c)}\left(1-p_{w}^{(c)}\right)+p_{w}^{(c)}<i^{(c)}\right)\\
\frac{p_{h}^{(c)}}{p_{w}^{(c)}}(i^{(c)}-p_{x}^{(c)}(1-p_{w}^{(c)}))+p_{x}^{(c)}(1-p_{h}^{(c)})\;\;\;\;\left(\text{others}\right)
\end{cases},
$}
\label{eq:ga_appendix}
\end{equation}
which amplifies image values while avoiding overflow from $[0,1]$ without clipping. The inverse function is
\begin{equation}
\resizebox{1.0\hsize}{!}{$
f_{GA}^{(c)-1}(I^{(c)}; P_{GA}^{(c)})=\begin{cases}
\frac{1-p_{w}^{(c)}}{1-p_{h}^{(c)}}i^{(c)}\;\;\;\;\left(\text{if}\,i^{(c)}<p_{x}^{(c)}\left(1-p_{h}^{(c)}\right),\,i^{(c)}\in I^{(c)}\right)\\
\frac{1-p_{w}^{(c)}}{1-p_{h}^{(c)}}i^{(c)}+\frac{p_{w}^{(c)}-p_{h}^{(c)}}{1-p_{h}^{(c)}}\;\;\;\,\left(\text{if}\,p_{x}^{(c)}\left(1-p_{h}^{(c)}\right)+p_{h}^{(c)}<i^{(c)}\right)\\
\frac{p_{w}^{(c)}}{p_{h}^{(c)}}(i^{(c)}-p_{x}^{(c)}(1-p_{h}^{(c)}))+p_{x}^{(c)}(1-p_{w}^{(c)})\;\;\;\;\left(\text{others}\right)
\end{cases}.
$}
\label{eq:ga_inv_appendix}
\end{equation}

\subsubsection*{Gamma Tone Mapping}

The gamma tone mapping $f_{GM}$ consists of three controllable parameters $P_{GM}^{(c)}=\{{p_{g1}^{(c)}, p_{g2}^{(c)}, p_{k}^{(c)}}\}$ for each color channel $c$ as follows:
\begin{equation}
f_{GM}^{(c)}(I^{(c)}; P_{GM}^{(c)})={I^{(c)}}^{{\frac{1}{p_{g1}^{(c)}}\cdot\frac{1-\left(1-p_{g2}^{(c)}\right){I^{(c)}}^{\frac{1}{p_{g1}^{(c)}}}}{1-\left(1-p_{g2}^{(c)}\right){p_{k}^{(c)}}^{\frac{1}{p_{g1}^{(c)}}}}}}.
\label{eq:gm_appendix}
\end{equation}

Because this function is not invertible, we approximate its inverse function using a 4-D lookup table $T$ that adds one dimension for the input pixel to these three parameters. The 4-D lookup table is linearly interpolated to get better approximation as follows:
\begin{equation}
\resizebox{0.908\hsize}{!}{$
f_{GM}^{(c)-1}(i^{(c)}; P_{GM}^{(c)})=interpolate(T,(i^{(c)},p_{g1}^{(c)},p_{g2}^{(c)},p_{k}^{(c)})).
\label{eq:lut_appendix}
$}
\end{equation}
The more details about the lookup table-based approximation is written in Section \ref{sec:lut}.

\subsubsection*{Contrast Stretcher}

The same function as the eq. \ref{eq:ga_appendix} is used for the contrast stretcher $f_{CS}$.

\subsubsection*{Inverse Tone Mapping}

The inverse tone mapping function $f_{IT}$ consists of three controllable parameters $P_{IT}^{(c)}=\{{p_{g3}^{(c)}, p_{g4}^{(c)}, p_{k2}^{(c)}}\}$ for each color channel $c$ as follows:
\begin{equation}
f_{IT}(I^{(c)}; P_{IT}^{(c)})={I^{(c)}}^{p_{g3}^{(c)}\cdot\frac{1+p_{g4}^{(c)}(I^{(c)}+1)^{p_{g3}^{(c)}}}{1+p_{g4}^{(c)}(p_{k2}^{(c)}+1)^{p_{g3}^{(c)}}}}.
\label{eq:it_appendix}
\end{equation}

Because this function is not invertible the samely with the $f_{GM}$, its inverse function is also approximated using a 4-D lookup table.

\begin{table*}[t]
\centering
\caption{Hyperparameter setting for the training phase of $f_{ISP}$. We change the hyperparameter setting according to the amount of training data and whether denoising is needed or not.}

\label{tab:hyp_f}
\scalebox{0.97}{
\begin{tabular}{cc|c|c|c}
\hline 
 & \multicolumn{3}{c|}{w/o denoising}& w/ denoising\\
\hline 
\# training data & $\leq 100$ & $\leq 1000$ & $>1000$  &$\leq 1000$\\
\hline 
\hline 
learning rate & 2e-5& 2e-5& 1e-4  &5e-5\\
batch size & 4 & 4 & 16  &8\\
iterations & 15,000& 15,000& 30,000 &70,000\\
loss $L_{ISP}$ & \multicolumn{3}{c|}{$L_{mse}+0.1L_{vgg}$} & $L_1(X,Y)+\sum_{k=\{16,32,64\}}L_{1}(A_k(X),A_k(Y))$\\
 optimizer& \multicolumn{4}{c}{AdamW}\\
  lerning rate schedule& \multicolumn{4}{c}{Cosine annealing whose minimal learning rate is 0.001 times of the max learning rate}\\
augmentation & \multicolumn{4}{c}{random flip and crop}\\
memorized ISP parameter sets $N$& \multicolumn{4}{c}{5000}\\
training image size & \multicolumn{4}{c}{$640\times480$}\\

\hline
\end{tabular}
}
\end{table*}

\begin{table*}[t]
\centering
\caption{Hyperparameter setting for the semi-supervised learning of the target model $F_{ISP}$.}

\label{tab:hyp_F}
\scalebox{0.99}{
\begin{tabular}{cc|c|c|c}
\hline 
 & \multicolumn{3}{c|}{w/o denoising}& w/ denoising\\
\hline 
\# training data & $\leq 100$ & $\leq 1000$ & $> 1000$ & $\leq 1000$ \\
\hline 
\hline 
learning rate & \multicolumn{4}{c}{1e-4} \\
batch size & $N_r=2, N_p=18$ & $N_r=4, N_p=16$ & $N_r=6, N_p=14$  & $N_r=4, N_p=16$\\
threshold $\beta$ & 1.0 & 1.2 & 1.0 & 1.2\\
iterations &  \multicolumn{4}{c}{40,000}\\
loss $L$ & \multicolumn{3}{c|}{$L_{mse}$} & $L_{mse}+L_{ssim}$\\
 optimizer& \multicolumn{4}{c}{AdamW} \\
 lerning rate schedule& \multicolumn{4}{c}{Cosine annealing whose minimal learning rate is 0.001 times of the max learning rate} \\ 
 augmentation & \multicolumn{4}{c}{random flip and crop}\\
training image size & \multicolumn{3}{c|}{$640\times480$} & $128\times128$\\
\hline
\end{tabular}
}
\end{table*}

\subsection{Lookup table-based approximation of non-invertible ISP functions}
\label{sec:lut}
The $f_{GM}$ and $f_{IT}$ are not invertible. Here we describe in detail how to approximate the inverse functions using a lookup table in such cases. First, as shown in the pseudocode in Algorithm \ref{alg:lookup} (1), a lookup table is obtained with an evenly spaced grid, taking the ISP parameters and the input images as input. Then, as shown in Algorithm \ref{alg:lookup} (2), the inverse functions are approximated using this lookup table. Additionally, the lookup table is complemented with linear interpolation to improve the accuracy of the approximation. For example, by setting the grid sizes $d_{g1}$, $d_{g2}$, $d_{k}$, $d_{i}$, $d_{o}$ to 60, 30, 30, 10000, and 512 respectively, and creating a 4-dimensional lookup table $T \in \mathbb{R}^{50\times30\times30\times512}$, the error for $|f_{GM}(f_{GM}^{-1}(I; P_{GM}); P_{GM}) - I|$ becomes 1.1e-5 on average and 6.5e-5 on maximum, resulting in a sufficiently accurate inverse function. By using RegularGridInterpolator implemented with CuPy's GPU parallel computation, the inference speed is about 1 ms, which does not become a bottleneck of the speed.

\subsection{Hyperparameters and Training Settings}

Our method consists of three training phases. The first phase involves training $f_{ISP}$ to generate pseudo-data. The second phase is semi-supervised training of the target model $F_{ISP}$. The final phase is the fine-tuning of $F_{ISP}$ using only supervised data. As mentioned in the main paper, we use the same hyperparameters as much as possible, but we adjust the hyperparameters for the first two training phases based on the amount of training data and the task. The hyperparameters and training setting for $f_{ISP}$ and $F_{ISP}$ are listed in Table \ref{tab:hyp_f} and Table \ref{tab:hyp_F} During fine-tuning, we use the hyperparameters from the original paper of the target model for supervised training. Since some training has already been done, we halve the number of iterations compared to the original paper. 

For the LoLv1 IE task that requires denoising, the input images of the supervised data are pre-denoised using KBNet \cite{zhang2023kbnet}  trained on the SenseNoise dataset \cite{zhang2022idr}, and then $f_{ISP}$ is trained on the data with reduced noise. Since it was found that KBNet is not good at denoising dark images, the images are first normalized to a range of 0 to 1, then gamma corrected with $y=x^{1/2.2}$, denoised with KBNet, and finally degamma corrected and denormalized to save the images with their original brightness. In the semi-supervised learning of the target model $F_{ISP}$, the amount of noise added to the pseudo-RAW images is randomized with $\sigma_s^2=\mathcal{U}(0, 0.01)$ and $\sigma_r^2=\mathcal{U}(0, 0.0002)$. 

\begin{table*}[t]
\centering
\caption{Additional ablation studies about sRGB-to-RAW including FiveK general ISP task and three target models. To fairly evaluate the sRGB-to-RAW methods themselves, we use the proposed updating sRGB and loss value-based data filtering for all methods. }

\scalebox{0.91}{

\label{tab:srgb2raw_appendix}

\begin{tabular}{lccccc|cccc}
\hline 
 &  & \multicolumn{4}{c}{FiveK tone mapping}& \multicolumn{4}{c}{FiveK general ISP}\\
\hline 
 & \# images & \multicolumn{2}{c}{100}  & \multicolumn{2}{c}{4500 (full)} & \multicolumn{2}{c}{100}  & \multicolumn{2}{c}{4500 (full)}\\
\hline 
sRGB-to-RAW & target model & PSNR & SSIM & PSNR & SSIM & PSNR & SSIM & PSNR & SSIM\\
\hline 
\hline 
UPI \cite{brooks2019unprocessing} & \multirow{6}{*}{NAFNet \cite{chen2022simple}} & 23.64& 0.909& 24.64& 0.914&21.20&0.870&22.44&0.879\\
 CycleISP \cite{zamir2020cycleisp}& & 23.35& 0.908& 24.45&0.913&21.15&0.867&22.54&0.877\\
 SRISP \cite{otsuka2023self}& & 23.67& \textbf{0.911}& 24.46&0.913&21.41&0.870&22.35&0.878\\
 RAWDiffusion \cite{reinders2024raw}& & 23.12& 0.905& 23.45&0.914&20.68&0.859&22.45&\textbf{0.880}\\
 ReRAW& & 23.35& 0.908& 24.86&\textbf{0.918}&20.84&0.860&22.50&0.875\\
ours &  & \textbf{23.81}& 0.907& \textbf{24.87}& \textbf{0.918}&\textbf{21.48}&\textbf{0.872}&\textbf{22.55}&\textbf{0.880}\\
\hline 
UPI \cite{brooks2019unprocessing} & \multirow{6}{*}{SepLUT \cite{yang2022seplut}} & 23.55& 0.899& 25.45& 0.921&21.26&0.864&23.03&0.895\\
 CycleISP \cite{zamir2020cycleisp}& & 23.69& 0.902& 25.35&0.920&21.38&0.862&23.07&0.894\\
 SRISP \cite{otsuka2023self}& & 23.79& \textbf{0.905}& 25.44&\textbf{0.922}&21.27&0.863&23.02&0.892\\
 RAWDiffusion \cite{reinders2024raw}& & 23.53& 0.900& 25.40&0.920&21.18&0.852&23.08&0.895\\
 ReRAW& & 23.80& 0.904& 25.40&\textbf{0.922}&21.01&0.850&23.02&0.895\\
ours &  & \textbf{23.98}& \textbf{0.905}& \textbf{25.45}& \textbf{0.922}&\textbf{21.66}&\textbf{0.873}&\textbf{23.18}&\textbf{0.897}\\
\hline 
UPI \cite{brooks2019unprocessing} & \multirow{6}{*}{PQDynamicISP \cite{yoshimura2024pqdynamicisp}} & 23.40& 0.904& 25.67& 0.932&21.60&0.879&23.61&0.910\\
 CycleISP \cite{zamir2020cycleisp}& & 23.32& 0.900& 25.55&0.929&21.57&0.875&23.46&0.911\\
 SRISP \cite{otsuka2023self}& & 23.39& 0.899& 25.62&0.932&20.99&0.867&\textbf{23.72}&0.911\\
 RAWDiffusion \cite{reinders2024raw}& & 23.21& 0.897& 25.63&0.932&20.98&0.867&23.40&0.909\\
 ReRAW& & 23.82& 0.901& 25.57&0.930&21.54&0.881&23.40&0.907\\
ours &  & \textbf{24.12}& \textbf{0.910}& \textbf{25.73}& \textbf{0.934}&\textbf{21.94}&\textbf{0.885}&23.61&\textbf{0.913}\\
\hline 
\end{tabular}
}
\end{table*}

\begin{table}
\centering
\caption{Evaluation on the FiveK IE task with 100, 500, and 4500 paired training images. In SemiIE (ours), sRGB images from COCO dataset are used as normal sRGB images.}
\vspace{-2mm}
\scalebox{0.74}{

\label{tab:fivek_srgb}

\begin{tabular}{lcccccc}
\hline 
 \# images & \multicolumn{2}{c}{100} & \multicolumn{2}{c}{500} & \multicolumn{2}{c}{4500 (full)}   
\\
 \hline 
 & PSNR & SSIM & PSNR & SSIM & PSNR & SSIM  \\
\hline 
\hline 
HDRNet \cite{gharbi2017deep} & - & - & - & - & 24.66 & 0.915 
 \\
3D LUT \cite{zeng2020learning} & - & - & - & - & 25.21 & 0.922 
 \\
CSRNet \cite{he2020conditional} & - & - & - & - & 25.17 & 0.921\\
AdaInt \cite{yang2022adaint} & - & - & - & - & 25.49 & 0.926 
 \\
NamedCurves \cite{serrano2024namedcurves} & - & - & - & - & 25.59 & 0.936 \\
LUTwithBGrid \cite{kim2024image} & - & - & - & - & 25.66 & 0.930 
 \\
\hline 
NAFNet-s \cite{chen2022simple} & 21.30 & 0.827 & 22.41 & 0.854 & 24.52 & \textbf{0.912}\\
 \quad+ SemiIE (ours)& \textbf{22.73}& \textbf{0.855}& \textbf{23.93}& \textbf{0.898}& \textbf{24.73}&  0.904\\
SepLUT \cite{yang2022seplut} & 21.54& 0.858& 22.38& 0.871& \textbf{25.47} & \textbf{0.921} \\
 \quad+ SemiIE (ours)& \textbf{23.04}& \textbf{0.895}& \textbf{24.28}& \textbf{0.908}& 25.30& 0.919 \\
PQDynamicISP \cite{yoshimura2024pqdynamicisp} & 22.63 & 0.893& 23.39& \textbf{0.896}& \textbf{25.53}& \textbf{0.928} \\
 \quad+ SemiIE (ours)& \textbf{23.59}& \textbf{0.905}& \textbf{23.74}& 0.891& 25.46 & \textbf{0.928} \\
\hline 
\end{tabular}
}
\vspace{-2mm}
\end{table}

\section{Additional Experiments}
\subsection{More Ablation Studies}
\begin{figure*}
    \centering
    \renewcommand{\arraystretch}{0.2}
    \setlength{\tabcolsep}{0.5pt}
    \begin{tabular}{cc}
    \includegraphics[width=0.42\textwidth]{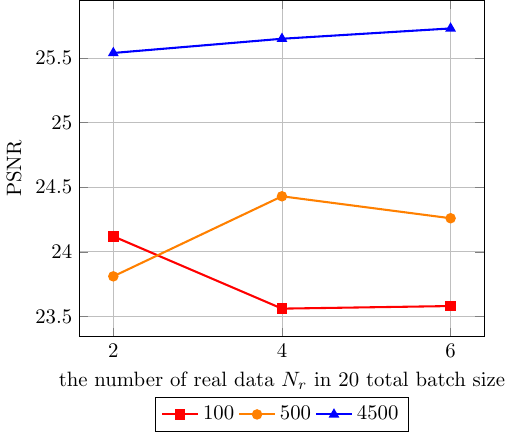} &
    \includegraphics[width=0.4\textwidth]{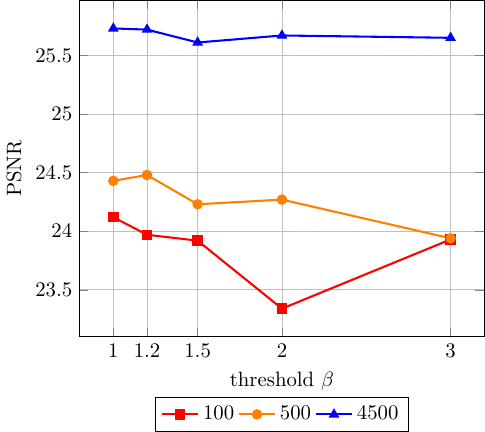}
    \end{tabular}
    \vspace{0mm}
    
    \caption{The left figure shows the results of changing the ratio of real data $N_r$ to pseudo-data $N_p$ batch sizes during semi-supervised learning in the FiveK tone mapping task. The right figure shows the results of changing the threshold $\beta$.}
    \vspace{0mm}
    \label{fig:ablation_hyper}
\end{figure*}

Additional ablation studies that cannot be included in the main paper are shown here. First, we present the results of an ablation study on the ratio of the real data batch size $N_r$ to the pseudo-data batch size $N_p$ during semi-supervised learning for the FiveK tone mapping task in Fig. \ref{fig:ablation_hyper}. The total batch size is fixed at 20, and only the ratio is varied. It is found that increasing the proportion of pseudo-data improves performance when there is a small amount of labeled data. On the other hand, when there are sufficient labeled data, it is better not to increase the ratio of pseudo-data too much.

Next, we conduct an ablation study on the value of the threshold $\beta$ during semi-supervised learning. Here, we use the optimal batch sizes for each amount of training data obtained above. Fig. \ref{fig:ablation_hyper} indicates that around $\beta=1$ is optimal regardless of the number of training samples if we choose the optimal batch size ratio.

Additionally, comparisons of the sRGB-to-RAW methods are conducted not only for the FiveK tone mapping task but also for the FiveK general ISP task. The results with three target models, NAFNet, SepLUT, and PQDynamicISP, are shown in Table \ref{tab:srgb2raw_appendix}. In the tone mapping task, white balance and color correction have already been applied to the RAW images, so the input data is in a relatively unified format. Therefore, the advantage of our one-to-many mapping sRGB-to-RAW over the deterministic sRGB-to-RAW methods is that it can account for variations in input data caused by incomplete color correction and white balance. In contrast, in the general ISP task, images taken with various cameras in various environments are input without metadata information, making it a more one-to-many mapping task. Therefore, the proposed method achieves significant performance improvements. It is also found that the proposed method works effectively with various types of target models.

\begin{table*}[t]
\centering
\caption{Evaluation on LoLv1, real LoLv2, and synthetic LoLv2 IE task. We report the metric after fixing global color using ground truth mean in (\;) following KinD \cite{zhang2019kindling}}
\vspace{-2mm}
\scalebox{0.99}{
\label{tab:lolv2}

\begin{tabular}{lcc|cc|cc}
 \hline 
 & \multicolumn{2}{c|}{LoLv1} & \multicolumn{2}{c|}{LoLv2} & \multicolumn{2}{c}{LoLv2-syn} \\
 & PSNR & SSIM & PSNR & SSIM 
& PSNR &SSIM 
\\
\hline 
\hline 
 LLFlow \cite{wang2022low}& 21.15 (25.19) & \;\;\;\;-\;\;\; \;(\;\;\;-\;\;\;)  & 17.43 (25.42)& \;\;\;\;-\;\;\; \;(\;\;\;-\;\;\;)& 24.81 (27.96)&\;\;\;\;-\;\;\; \;(\;\;\;-\;\;\;)  
\\
 SNR-Aware \cite{xu2022snr} & 24.61 (26.72) & 0.842 (0.851) & 21.48 (27.81)& 0.849 (0.871)& 24.14 (27.79)&0.842 (0.851) 
\\
 CIDNet \cite{yan2025hvi}& 23.81 (27.72) & 0.857 (0.876) & 24.11 (28.13)& 0.868 (0.892)& 25.67 (28.99)&0.930 (0.939)\\
\hline 
DRBN \cite{yang2020fidelity} & 16.29 (19.55)& 0.617 (0.746) & 20.29 \;(\;\;\;-\;\;\;)& 0.831 \;(\;\;\;-\;\;\;)& 23.22 \;(\;\;\;-\;\;\;)&0.927 \;(\;\;\;-\;\;\;)\\
SMSNet \cite{malik2023semi} & 21.91 \;(\;\;\;-\;\;\;)& 0.780 \;(\;\;\;-\;\;\;) & \;\;\;\;-\;\;\; \;(\;\;\;-\;\;\;)  
& \;\;\;\;-\;\;\; \;(\;\;\;-\;\;\;)  
& \;\;\;\;-\;\;\; \;(\;\;\;-\;\;\;)  
&\;\;\;\;-\;\;\; \;(\;\;\;-\;\;\;)  
\\
CRNet \cite{lee2023temporally} & \;\;\;\;-\;\;\; \;(24.01) & \;\;\;\;-\;\;\; \;(\;\;\;-\;\;\;)  & \;\;\;\;-\;\;\; \;(\;\;\;-\;\;\;)  
& \;\;\;\;-\;\;\; \;(\;\;\;-\;\;\;)  
& \;\;\;\;-\;\;\; \;(\;\;\;-\;\;\;)  
&\;\;\;\;-\;\;\; \;(\;\;\;-\;\;\;)  
\\
LMT-GP \cite{yu2024lmt} & 24.12 \;(\;\;\;-\;\;\;) & \;\;\;\;-\;\;\; \;(\;\;\;-\;\;\;)  & \;\;\;\;-\;\;\; \;(\;\;\;-\;\;\;)  & \;\;\;\;-\;\;\; \;(\;\;\;-\;\;\;)  & \;\;\;\;-\;\;\; \;(\;\;\;-\;\;\;)  &\;\;\;\;-\;\;\; \;(\;\;\;-\;\;\;)  \\
\hline 
LLFormer \cite{wang2023ultra} & 23.65 (25.76) & 0.816 (0.823) & 20.06 (26.20)& 0.792 (0.819)& 24.04 (28.01)&0.909 (0.927)\\
\quad+ SemiIE (ours) & \textbf{23.77 (26.12)}& \textbf{0.832 (0.841)} & \textbf{21.66 (28.08)}& \textbf{0.841 (0.868)}& \textbf{25.33 (28.79)}&\textbf{0.933 (0.942)}\\
Retinexformer \cite{Cai_2023_ICCV} & \textbf{23.89 (27.00)}& 0.835 (0.853) & 21.81 (28.16)& 0.837 (\textbf{0.878})& 25.89 (29.37)&0.934 (0.943)\\
\quad+ SemiIE (ours) &  23.78 (\textbf{27.00})& \textbf{0.837 (0.854)} & \textbf{22.34 (28.39)}& \textbf{0.844} (0.870)& \textbf{26.17 (29.40)}&\textbf{0.935 (0.944)}\\
\hline 
\end{tabular}
}
\vspace{-1mm}
\end{table*}

\subsection{More Benchmarking}
Additional experiments with FiveK \cite{bychkovsky2011learning}, LoLv2 real \cite{yang2021sparse}, and LoLv2 synthetic \cite{yang2021sparse} datasets are shown here.

\subsubsection*{FiveK IE task}
This task converts sRGB images from hardware ISPs into sRGB images retouched by expert C. The image size and train test splitting are identical to those for FiveK tone mapping and FiveK general ISP tasks.
The results of training with 100, 500, and 4500 data samples are shown in Table \ref{tab:fivek_srgb}. As with the FiveK tone mapping task and the FiveK general ISP task, the proposed method significantly improves accuracy when training data is limited. However, when there is a large amount of training data, the accuracy does not improve. This may be because the adaptation of the idea of one-to-many mapping sRGB-to-RAW for ISP to IE by adding an inverse tone mapping function to the ISP function might have been somewhat forced. Further improvements are desired for the IE task in the future.

\subsubsection*{LoLv1, LoLv2, and LoLv2-syn IE tasks}

The detailed results for the LoLv1, LoLv2, and LoLv2-syn IE tasks are shown in Table \ref{tab:lolv2}.

\section{Visual Comparison}

\begin{figure*}
\centering
\renewcommand{\arraystretch}{0.2}
\setlength{\tabcolsep}{0.5pt}
\scalebox{1.0}{
\begin{tabular}{cccc}
\includegraphics[width=0.245\textwidth]{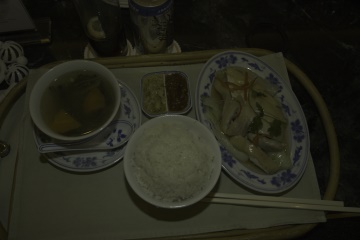}& \includegraphics[width=0.245\textwidth]{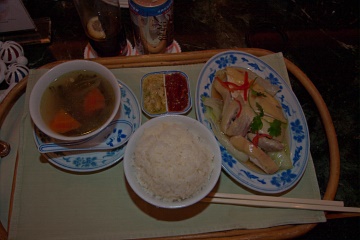}&\includegraphics[width=0.245\textwidth]{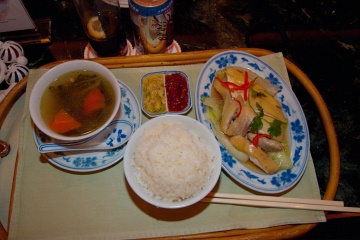}&
\includegraphics[width=0.245\textwidth]{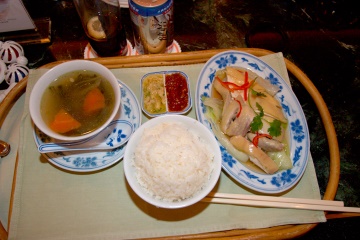}\\
\includegraphics[width=0.245\textwidth]{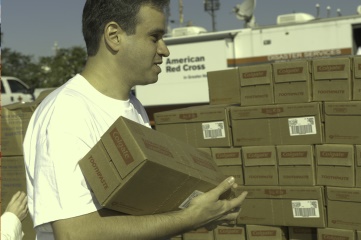}& \includegraphics[width=0.245\textwidth]{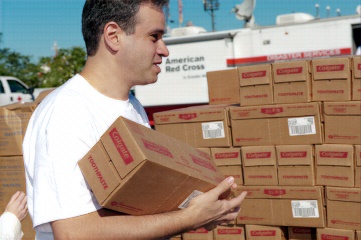}&\includegraphics[width=0.245\textwidth]{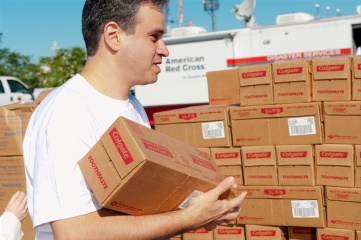}&
\includegraphics[width=0.245\textwidth]{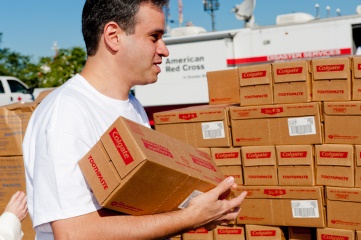}\\
 &  && \\
input& NAFNet&+ SemiISP (ours)& Ground truth\\
 &  && \\
  &  && \\
\includegraphics[width=0.245\textwidth]{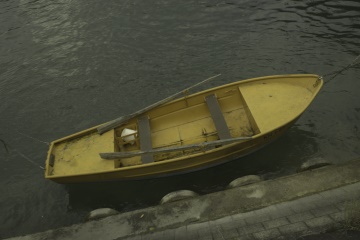}& \includegraphics[width=0.245\textwidth]{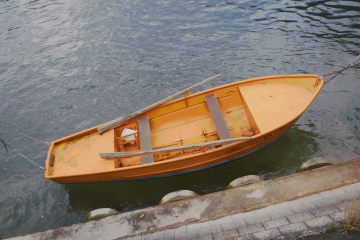}&\includegraphics[width=0.245\textwidth]{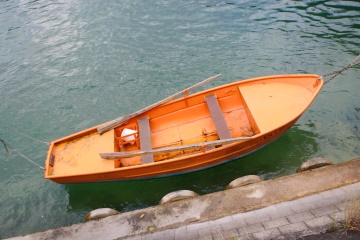}&
\includegraphics[width=0.245\textwidth]{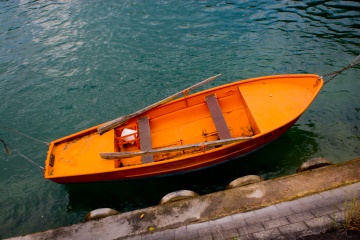}\\
\includegraphics[width=0.2\textwidth]{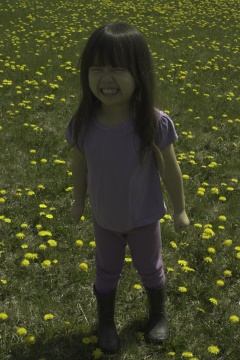}& \includegraphics[width=0.2\textwidth]{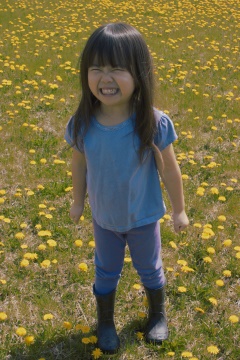}&\includegraphics[width=0.2\textwidth]{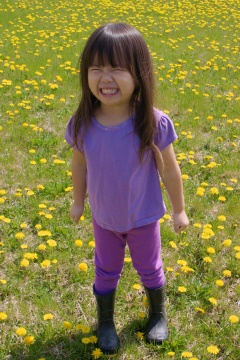}&
\includegraphics[width=0.2\textwidth]{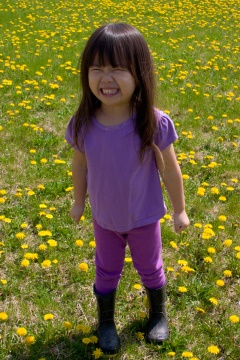}\\
 &  && \\
input& SepLUT &+ SemiISP (ours)& Ground truth\\
 &  && \\
  &  && \\
\includegraphics[width=0.245\textwidth]{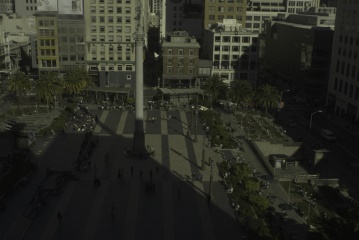}& \includegraphics[width=0.245\textwidth]{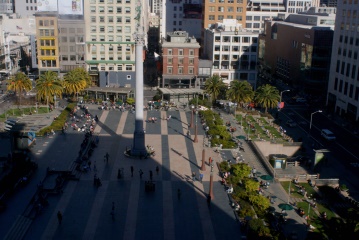}&\includegraphics[width=0.245\textwidth]{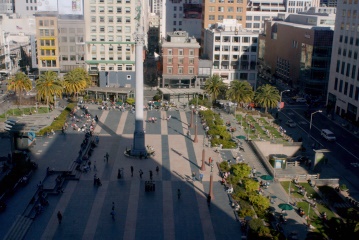}&
\includegraphics[width=0.245\textwidth]{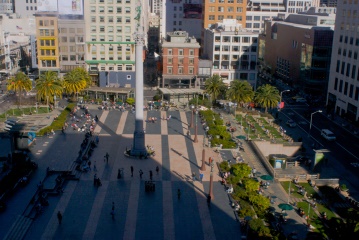}\\
\includegraphics[width=0.245\textwidth]{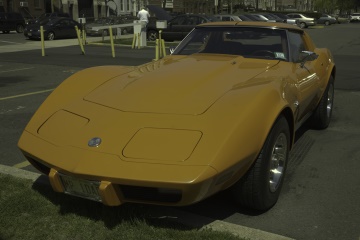}& \includegraphics[width=0.245\textwidth]{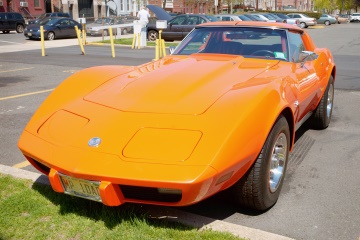}&\includegraphics[width=0.245\textwidth]{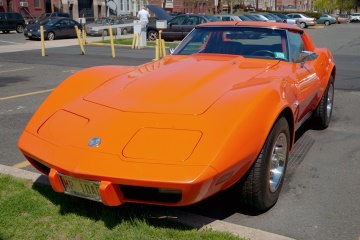}&
\includegraphics[width=0.245\textwidth]{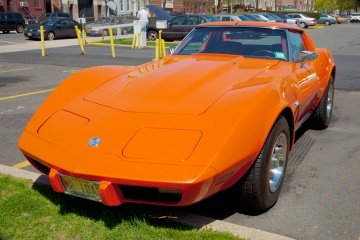}\\
 &  && \\
input& PQDynamicISP&+ SemiISP (ours)& Ground truth\\
\end{tabular}
}
\vspace{-2mm}

\caption{Visual comparison of training with only 100 training samples in the FiveK tone mapping task. The proposed SemiISP realizes ISPs that are stable and high-quality with just 100 training samples. }
\vspace{0mm}
\label{fig:viz_fivek_tone}
\end{figure*}

\begin{figure*}
\centering
\renewcommand{\arraystretch}{0.2}
\setlength{\tabcolsep}{0.5pt}
\scalebox{1.0}{
\begin{tabular}{cccc}
\includegraphics[width=0.245\textwidth]{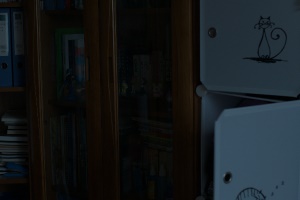}& \includegraphics[width=0.245\textwidth]{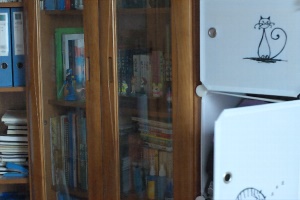}&\includegraphics[width=0.245\textwidth]{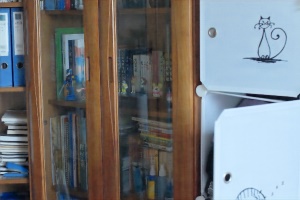}&
\includegraphics[width=0.245\textwidth]{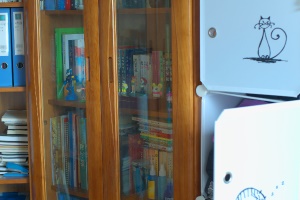}\\
\includegraphics[width=0.245\textwidth]{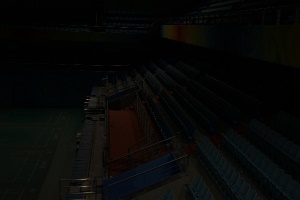}& \includegraphics[width=0.245\textwidth]{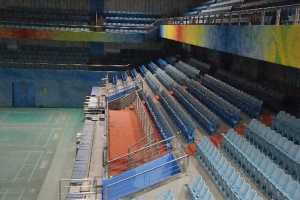}&\includegraphics[width=0.245\textwidth]{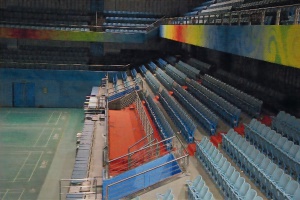}&
\includegraphics[width=0.245\textwidth]{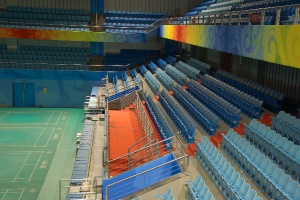}\\
 &  && \\
input& LLFormer &+ SemiIE (ours)& Ground truth\\
\end{tabular}
}
\vspace{0mm}

\caption{Visual comparison in the LoLv1 IE task.}
\vspace{0mm}
\label{fig:viz_lolv1}
\end{figure*}

\begin{figure*}
\centering
\renewcommand{\arraystretch}{0.2}
\setlength{\tabcolsep}{0.5pt}
\scalebox{1.0}{
\begin{tabular}{cccc}
\includegraphics[width=0.245\textwidth]{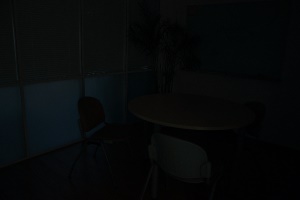}& \includegraphics[width=0.245\textwidth]{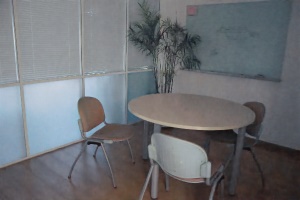}&\includegraphics[width=0.245\textwidth]{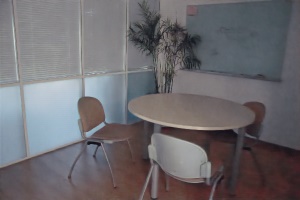}&
\includegraphics[width=0.245\textwidth]{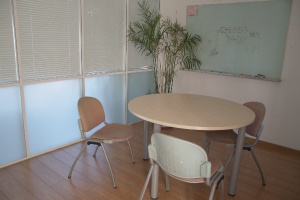}\\
\includegraphics[width=0.245\textwidth]{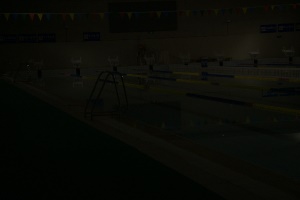}& \includegraphics[width=0.245\textwidth]{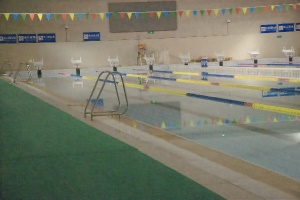}&\includegraphics[width=0.245\textwidth]{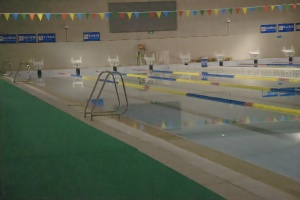}&
\includegraphics[width=0.245\textwidth]{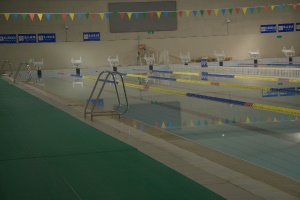}\\
 &  && \\
input& Retinexformer &+ SemiIE (ours)& Ground truth\\
\end{tabular}
}
\vspace{0mm}

\caption{Visual comparison in the LoLv2 IE task.}
\vspace{0mm}
\label{fig:viz_lolv2}
\end{figure*}

In this section, we compare the visual appearance of the predictions from trained target models. Table \ref{fig:viz_fivek_tone} is a visual comparison of training with only 100 training samples in the FiveK tone mapping task. The proposed SemiISP realizes ISPs that are stable and high-quality with just 100 training samples. Table \ref{fig:viz_lolv1} and Table \ref{fig:viz_lolv2} show the visual comparisons for the LoLv1 and LoLv2 IE tasks, indicating not only tone mapping quality but also denoising quality are improved by the proposed SemiIE.

\section{Limitation and Future Work}

Our method successfully improved the performance of the target models across various amounts of training data. However, it was necessary to adjust the hyperparameters especially the batch size depending on the amount of training data. Although it is a general problem that optimal hyperparameters differ depending on the amount of training data, methods that can be universally applied without adjusting hyperparameters based on the amount of training data are desirable in the future. Additionally, although we used the same hyperparameters regardless of the dataset if the training data size is the same, it might be possible to further improve performance by adjusting them. A mechanism that automatically adjusts the hyperparameters would also be desirable in the future.

In addition, the performance improvement in IE tasks using the proposed method is less than the performance improvement in ISP tasks. This may be because the adaptation of the idea of one-to-many mapping sRGB-to-RAW for ISP to IE by adding an inverse tone mapping function to the ISP function might have been somewhat forced. Further improvements are desired for the IE task in the future.

\begin{algorithm*}[tb] 
\caption{Pseudo-code of the lookup table-based inverse function approximation in numpy (np) style. We showcase an example for $f_{GM}$, but it can be applied to any functions.}\label{alg:lookup} 

\textbf{(1) Lookup table generation phase:} \\
\textbf{Input:} The known forward ISP function $f_{GM}$, the domain ranges of ISP parameters, $r_{p1}$, $r_{p2}$, and $r_{k}$, the number of grids for each parameter dimension of the lookup table, $d_{g1}$, $d_{g2}$, and $d_{k}$, the number of grid for the input and output image, $d_{i}$ and $d_{o}$. \\
\textbf{Output:} The inverse function of $f_{GM}.$

\SetAlgoLined
    \PyCode{} \\
    \PyCode{def generate($f_{GM}$, $r_{p1}$, $r_{p2}$, $r_{k}$, $d_{g1}$, $d_{g2}$, $d_{k}$, $d_{i}$, $d_{o}$):} \\
	\Indp
        \PyComment{prepare grids}\\
	\PyCode{g1s = np.linspace($r_{g1}$[0], $r_{g1}$[1], $d_{g1}$)} \\
        \PyCode{g2s = np.linspace($r_{g2}$[0], $r_{g2}$[1], $d_{g2}$)} \\
        \PyCode{ks = np.linspace($r_{k}$[0], $r_{k}$[1], $d_{k}$)} \\
        \PyCode{inps = np.linspace(0, 1, $d_{i}$)} \\
        \PyCode{outs = np.linspace(0, 1, $d_{o}$)} \\
        \PyCode{g1\_grid, g2\_grid, k\_grid, inp\_grid = np.meshgrid(g1s, g2s, ks, inps, indexing='ij', sparse=True)} \\
	\PyCode{} \\

        \PyComment{calculate the output for all combinations of parameters and input values}\\
        \PyCode{out\_grid = $f_{GM}$(inp\_grid; \{g1\_grid, g2\_grid, k\_grid\})} \\
        \PyCode{} \\

        \PyComment{create a LUT with evenly spaced grid} \\
        \PyCode{lut\_values = []} \\
        \PyCode{for i in range($d_{o}$):} \\
            \Indp
            \PyCode{nearest\_indices = np.argmin(np.abs(out\_grid - outs[i]), axis=-1)} \\
            \PyCode{nearest\_inp\_values = np.take\_along\_axis(inp\_grid, np.expand\_dims(nearest\_indices, axis=-1), axis=-1)} \\
            \PyCode{lut\_values.append(nearest\_inp\_values)} \\
            \Indm
        \PyCode{lut\_values = np.concatenate(lut\_values, axis=-1)} \\
        \PyCode{} \\

        \PyComment{create the lookup table function with linear interpolation}\\
        \PyComment{(g1s, g2s, ks, outs): grid information of LUT, lut\_values: the values of LUT}\\
        \PyCode{$f_{GM}^{-1}$ = scipy.interpolate.RegularGridInterpolator((g1s, g2s, ks, outs), lut\_values, method="linear")} \\ 
        
        \PyCode{return $f_{GM}^{-1}$} \\
        \PyCode{} \\
        \Indm

\textbf{(2) Apply the $f_{GM}^{-1}$ based on lookup table:} \\
\textbf{Input:} Input images: $I$, ISP parameters: $p_{g1}$, $p_{g2}$, and $p_{k}$, the inverse function: $f_{GM}^{-1}$. \\
\textbf{Output:} The output of $f_{GM}^{-1}.$

\SetAlgoLined
    \PyCode{} \\
    \PyCode{def apply($I$, $p_{g1}$, $p_{g2}$, $p_{k}$, $f_{GM}^{-1}$):} \\
	\Indp

        \PyComment{broadcast the ISP parameters}\\
        \PyCode{ones = np.ones\_like($I$)} \\
        \PyCode{$p_{g1}$, $p_{g2}$, $p_{k}$ = ones*$p_{g1}$, ones*$p_{g2}$, ones*$p_{k}$} \\

        \PyCode{input = np.concatenate([$p_{g1}$.reshape(-1, 1), $p_{g2}$.reshape(-1, 1), $p_{k}$.reshape(-1, 1), $I$.reshape(-1, 1)], axis=-1)} \\

        \PyComment{(apply the inverse function}\\
        \PyCode{out = $f_{GM}^{-1}$(input)} \\

        \PyCode{return out.reshape(*$I$.shape)} \\
        
        \Indm
\end{algorithm*}



\end{document}